%% file: bigdata.tex
\documentclass[lettersize,journal]{IEEEtran}
\usepackage[colorlinks,urlcolor=blue,linkcolor=blue,citecolor=blue]{hyperref}

\usepackage{color,array}

\usepackage{graphicx}
\PassOptionsToPackage{table}{xcolor}

\usepackage{amsmath,amsfonts}
\usepackage{array}
\usepackage{textcomp}
\usepackage{stfloats}
\usepackage{url}
\usepackage{cite}
\usepackage{verbatim}
\usepackage{graphicx}

\usepackage{times}
\usepackage{soul}
\usepackage{url}
\usepackage[utf8]{inputenc}
\usepackage[small]{caption}
\usepackage{amsmath}
\usepackage{amsthm}
\usepackage{booktabs}
\usepackage{algorithm}
\usepackage[switch]{lineno}
\usepackage{orcidlink}
\usepackage{amssymb}

\usepackage{mathtools} 
\usepackage{dsfont} 
\usepackage[noend]{algpseudocode}
\usepackage{multirow}
\usepackage{adjustbox}
\usepackage{makecell}
\usepackage{pifont}

\usepackage{subcaption}  
\usepackage{float}  

\newcommand{\major}[1]{\textcolor{black}{#1}}
\newcommand{\carlo}[1]{\textcolor{black}{#1}}

\newcommand{\tool}{{\textsc{{PUF}}}}

\begin{document}

\title{Federated Unlearning Made Practical: Seamless Integration via Negated Pseudo-Gradients}

\author{Alessio~Mora \orcidlink{0000-0001-8161-1070},
        ~Carlo~Mazzocca \orcidlink{0000-0001-8949-2221},
        ~Rebecca~Montanari~\orcidlink{0000-0002-3687-0361},
        ~Paolo~Bellavista \orcidlink{0000-0003-0992-7948}
\thanks{}
\thanks{Alessio Mora, Rebecca Montanari, and Paolo Bellavista are with the Department of Computer Science and Engineering, University of Bologna, Bologna, Italy (e-mail: \{name.surname\}@unibo.it).\\
Carlo Mazzocca is with the Department of Information and Electrical Engineering and Applied Mathematics, University of Salerno, Fisciano, Italy (email: cmazzocca@unisa.it).
}%
}

\markboth{Journal of \LaTeX\ Class Files,~Vol.~14, No.~8, August~2021}%
{Shell \MakeLowercase{\textit{et al.}}: A Sample Article Using IEEEtran.cls for IEEE Journals}



\maketitle

\input{sections/abstract}


\input{sections/introduction_anonymous}
\input{sections/related}

\input{sections/background}

\input{sections/method}

\input{sections/results}

\input{sections/conclusion}

\bibliographystyle{IEEEtran}
\bibliography{bib}

\input{sections/appendix}

\end{document}

%% file: sections/abstract.tex
\begin{abstract}
The \emph{right to be forgotten} is a fundamental principle of privacy-preserving regulations and extends to Machine Learning (ML) paradigms such as Federated Learning (FL). While FL enhances privacy by enabling collaborative model training without sharing private data, trained models still retain the influence of training data. Federated Unlearning (FU) methods recently proposed often rely on impractical assumptions for real-world FL deployments, such as storing client update histories or requiring access to a publicly available dataset. To address these constraints, this paper introduces a novel method that leverages negated Pseudo-gradients Updates for Federated Unlearning (\tool{}). Our approach only uses standard client model updates, which are employed during regular FL rounds, and interprets them as pseudo-gradients. When a client needs to be forgotten, we apply the negation of their pseudo-gradients, appropriately scaled, to the global model. Unlike state-of-the-art mechanisms, \tool{} seamlessly integrates with FL workflows, incurs no additional computational and communication overhead beyond standard FL rounds, and supports concurrent unlearning requests. We extensively evaluated the proposed method on two well-known benchmark image classification datasets (CIFAR-10 and CIFAR-100) and a real-world medical imaging dataset for segmentation (ProstateMRI), using three different neural architectures: two residual networks and a vision transformer. The experimental results across various settings demonstrate that \tool{} achieves state-of-the-art forgetting effectiveness and recovery time, without relying on any additional assumptions.
\end{abstract}

%% file: sections/introduction_anonymous.tex
\section{Introduction}

\IEEEPARstart{I}n today's digital landscape, privacy has become a major concern, as reflected by the emergence of robust regulatory frameworks worldwide \cite{arshad_survey}. The European Union (EU) has consistently emphasized the importance of protecting personal data, exemplified by the introduction of the General Data Protection Regulation (GDPR) in 2016 \cite{GDPR2016a}. Most recently, in May 2024, the EU enacted Regulation 2024/1183 \cite{eu2024regulation1183}, establishing the European Digital Identity Framework that empowers individuals with fine-grained control over their information.
One of the key rights of these regulations is the \emph{right to be forgotten}, which allows individuals to request the deletion of their previously shared data. Similar rights are central to other major privacy laws worldwide, \carlo{such as} the California Consumer Privacy Act (CCPA) \cite{californiaccpa} \carlo{where} the \emph{right to delete} grants California residents the on-demand removal of personal data \carlo{held by} businesses. 

Individuals should retain the ability to exercise their privacy rights even in Machine Learning (ML), \carlo{including withdrawing} their contributions from a trained model due to security or privacy concerns \cite{cao2015towards}. \carlo{This} can be achieved through Machine Unlearning (MU) \cite{shaik2024taxonomy, liu2024model, chundawat2023can}, an emerging paradigm designed to selectively erase the influence of specific training data from a model by post-processing the trained model. 

\input{figures/radar_cifar100_nonIID_1e}

The \emph{right to be forgotten} should \carlo{also} extend to privacy-preserving ML paradigms such as Federated Learning (FL) \cite{mcmahan2017communication, bonawitz2019towards}, \carlo{which} enables the collaborative training of a global model across multiple clients without requiring them to share their private data with a central server. In FL, clients train local models using their on-premises data and only share updates (e.g., weight differences) \cite{bellavista2021decentralised}. Despite offering enhanced privacy, the global model inevitably reflects the influence of client training data \cite{pmlr-v162-marfoq22a}. 

However, the intrinsic characteristics of FL, such as its non-deterministic and iterative training process, make traditional MU techniques less effective in this context \cite{romandini2024federated}. This has led to the novel concept of Federated Unlearning (FU) \cite{romandini2024federated,zhao2023survey}, which encompasses unlearning methods tailored for FL settings. The goal of an FU algorithm depends on what we are trying to remove from the global model, which could be either some samples, all the contributions of a client, or all samples belonging to a class. In this work,  we address \emph{client unlearning}, where a client seeks to remove all traces of its previously shared data from the global model. 
\input{tables/related}

An effective mechanism should achieve unlearning while maintaining the overall performance of the global model. Although \carlo{some} FU techniques have started to be proposed, most of them rely on hard assumptions as they usually require maintaining additional information, such as historical updates \cite{federaser,liu2022right,wu2022unlearningdistillation} or stateful clients \cite{halimi2022federated}. Moreover, \carlo{these works fail to address} scenarios where multiple unlearning requests occur in the same time window. This condition cannot be overlooked, as clients should be allowed to remove their contributions regardless of others' willingness.

To fill this gap, this paper presents a novel method that leverages negated Pseudo-gradients Updates for Federated Unlearning (\tool{}). Our approach ensures seamless integration into existing FL frameworks as it enables unlearning without modifying the traditional Federated Averaging (FedAvg) \cite{mcmahan2017communication} training protocol. Unlike most previous \carlo{works, \tool{} eliminates the need for} storing historical information (e.g., historical updates), \carlo{accessing} proxy data, \carlo{maintaining} stateful clients, or \carlo{incurring} impractical computational requirements. Since \tool{} can be seamlessly integrated into standard FedAvg, it also explicitly addresses multiple unlearning requests that concurrently occur. The client who wants to be forgotten sends an unlearning request to the server and performs \carlo{one final} round of FL. The server \carlo{then adjust}s its model updates by flipping and scaling them according to an unlearning rate. 

\carlo{To demonstrate the broad applicability of our approach, we implemented \tool{} and extensively evaluated its performance on two benchmark image classification datasets (CIFAR-10, CIFAR-100) and a real-world medical imaging dataset (ProstateMRI) for segmentation. The evaluation involved two residual networks and a vision transformer. The reported results clearly show the efficiency and effectiveness of \tool{} in removing client contributions to the global model across a variety of settings. Figure \ref{fig:radar_intro} visually compares the performance of \tool{} against \major{five} state-of-the-art mechanisms.} \major{It is worth noting that \tool{} can also be extended to sample unlearning with minimal modification.}

\smallskip
\noindent\textbf{Contributions.} The main contributions of this work can be summarized as follows:

\begin{itemize} \item We introduce \tool{}, a novel approach that enables FU without requiring any modifications or additional overhead to the original \carlo{FedAvg} algorithm.
\item We evaluate \tool{} across different federated settings and compare its performance to two state-of-the-art baselines for client unlearning. We show that \tool{} is more effective in removing the contribution of the target clients and more efficient in recovering the expected performance.
\item We open-source our code to support further related research in the community. The \tool{} code is available for the research community at: \url{https://anonymous.4open.science/r/puf_unlearning/}.
\end{itemize}

\smallskip
\noindent\textbf{Organization.} The remainder of this work is organized as follows. Section \ref{sec:background} provides the background on FL and FU. Section \ref{sec:related} reviews the main related work in the field. Section \ref{sec:approach} presents \tool{}, while Section \ref{sec:experimental_results} evaluates its performance and Section \ref{sec:conclusion} concludes the paper.

%% file: figures/radar_cifar100_nonIID_1e.tex
\begin{figure}[!t]
\centering
\includegraphics[width=0.85\columnwidth]{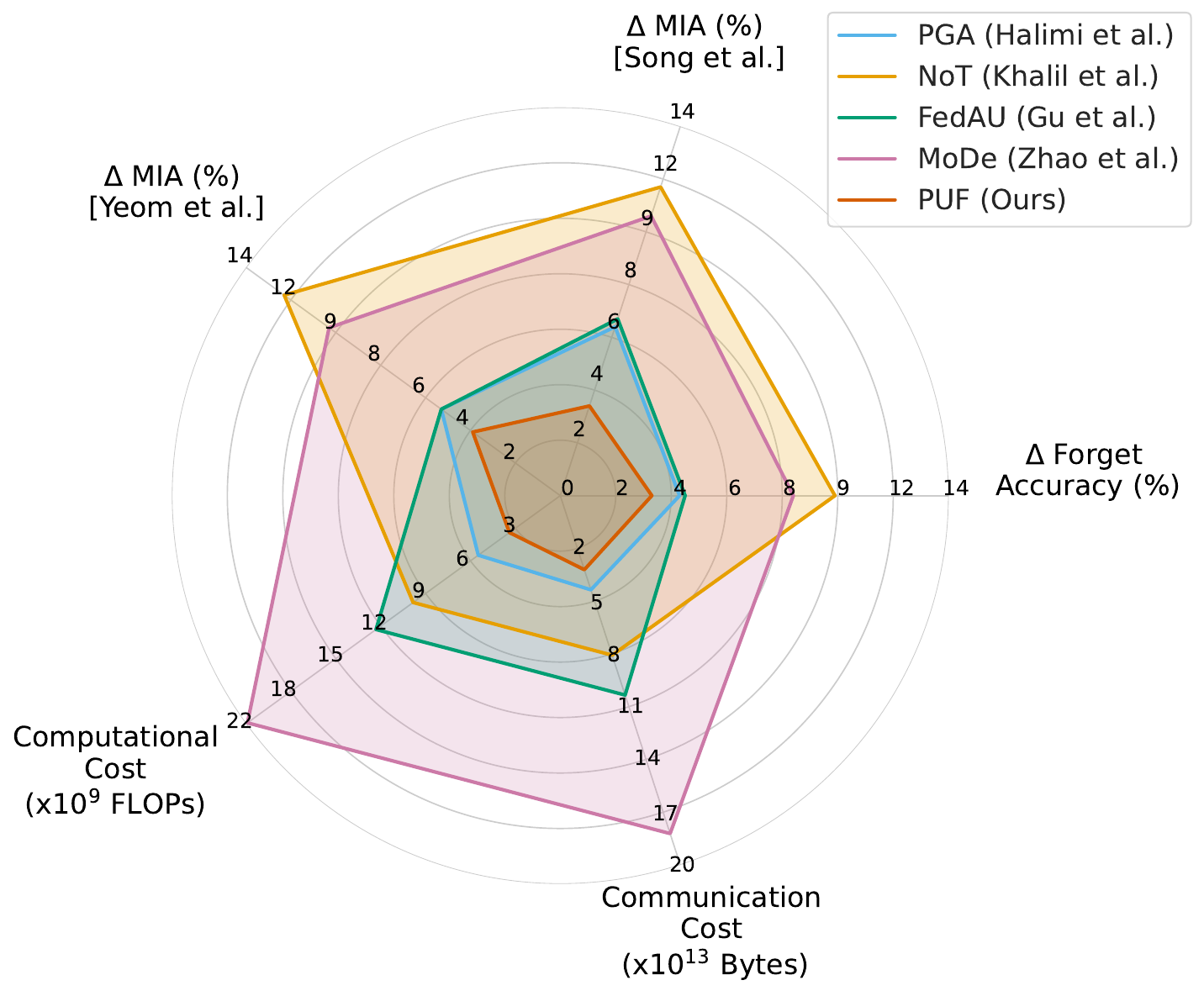}
\caption{\textbf{Performance comparison of PUF against state-of-the-art mechanisms} using ResNet-18 on CIFAR-100 (non-IID case) in a 10-client setup, where one client requests unlearning. The ideal FU algorithm should (1) minimize the difference with the gold standard \textit{retrained} model, across forgetting metrics such as Forget Accuracy and various MIAs (in Figure expressed as $\Delta$ Forget Accuracy and $\Delta$ MIAs), \major{(2) minimize computational and communication cost to recover model utility. A smaller polygon represents better unlearning performances. Experimental details are available in Section \ref{sec:experimental_results}, with full results across settings reported in Table \ref{table:full_results}.}}
\label{fig:radar_intro}
\end{figure} 
\captionsetup{labelfont={color=black}}

%% file: tables/related.tex
\begin{table*}[!t]
    \centering
    \adjustbox{max width=\textwidth}{
    \begin{tabular}{lcccccccc}
        \toprule
        \textbf{Work} & \textbf{\makecell{No Historical\\Information}} & \textbf{\makecell{No Full\\Participation}} & \textbf{\makecell{No Stateful\\Clients}} & \textbf{\makecell{Task\\Agnostic}} & \textbf{\makecell{No Auxiliary\\Data}} & \textbf{\makecell{Single-Round\\Unlearning}} & \textbf{\makecell{Simultaneous\\Unlearning}} & \textbf{\makecell{Selective\\Unlearning}}\\
        \midrule
        Liu et al. \cite{liu2022right} & $\times$ & $\times$ & $\checkmark$ & $\checkmark$ & $\checkmark$ & $\times$ & $\times$ & $\checkmark$\\
        FedEraser \cite{federaser}  & $\times$ & $\times$ & $\checkmark$ & $\checkmark$ & $\checkmark$ & $\times$ & $\times$ & $\checkmark$\\
        FUKD \major{\cite{wu2022unlearningdistillation, wu2024unlearning}} & $\times$ & $\checkmark$ & $\checkmark$ & $\times$ & $\times$ & $\checkmark$ & $\times$ & $\checkmark$\\ 
        PGA \cite{halimi2022federated} & $\checkmark$ & $\times$ & $\times$ & $\checkmark$ & $\checkmark$ & $\checkmark$ & $\times$ & $\checkmark$ \\ 
        \major{MoDe \cite{zhao2023federated}} & $\checkmark$ & $\checkmark$ & $\checkmark$ & $\times$ & $\checkmark$ & $\times$ & $\times$ & $\checkmark$\\ 
        \major{FedAU \cite{gu2024unlearning}} & $\checkmark$ & $\checkmark$ & $\checkmark$ & $\times$ & $\checkmark$ & $\checkmark$ & $\checkmark$ & $\checkmark$\\ 
        NoT \cite{khalil2025not} & $\checkmark$ & $\checkmark$ & $\checkmark$ & $\checkmark$ & $\checkmark$ & $\checkmark$ & $\checkmark$ & $\times$\\ 
        \rowcolor{lightgray!60} \textbf{\tool{}} \textbf{(ours)} & $\checkmark$ & $\checkmark$ & $\checkmark$ & $\checkmark$ & $\checkmark$ & $\checkmark$ & $\checkmark$ & $\checkmark$\\ 
        \bottomrule
    \end{tabular}
    }\caption{Comparison with other works performing client unlearning in federated settings.}\label{table:related}
\end{table*}

%% file: sections/related.tex
\section{Related Work}\label{sec:related}
Table \ref{table:related} presents a comparative analysis of \tool{} against existing approaches that target the client-unlearning case. The columns indicate whether each method does not rely on a given requirement ($\checkmark$) or does depend on it ($\times$). Our comparison highlights key aspects of FU methods, with a primary focus on whether an approach requires and leverages historical updates \cite{liu2022right,federaser,wu2022unlearningdistillation,wu2024unlearning}.
\major{Storing past client model updates raises serious scalability concerns\footnote{\major{To provide a concrete example, consider a ResNet-18 model with approximately 11 million model parameters (around 44 MB per update using 32-bit precision). Retaining updates from 1000 clients over 500 training rounds and across 10 learning tasks would require more than 220 TB of storage. Such a requirement is likely infeasible in practical FL deployments, especially as models, client counts, and training durations continue to grow \cite{bonawitz2019towards}.}}. Beyond storage overhead, this requirement introduces a critical limitation: the server must retain the ability to link each update to a specific client, which is fundamentally at odds with the ephemeral and privacy-preserving nature of FL model updates~\cite{kairouz2019advances}. Furthermore, because these updates must be pre-stored, the data to be removed must be specified in advance, e.g., the entire dataset of a target client. This limitation prevents such methods from supporting more flexible unlearning scenarios, such as when a client requests to remove only a subset of its data after training has already occurred.}

\major{In particular, FedEraser \cite{federaser} uses a multi-round
calibration mechanism, which iteratively corrects the historical updates for a set of past retained rounds. However, for
each retained round, the remaining clients in the federation must participate in the calibration step; hence, it requires clients being online and willing to participate in the calibration phase for several rounds. 
Wu et al. \cite{wu2022unlearningdistillation,wu2024unlearning} further assume access to auxiliary unlabeled data to mitigate performance degradation due to unlearning. However, to be effective, such a proxy dataset must be semantically aligned with the data used in the federation and balanced across classes  \cite{nayak2021effectiveness}. Furthermore, assuming the availability of a public dataset, and relying on it for unlearning in FL is often unrealistic in practice \cite{mora2024knowledge}.} 

Given these limitations, we \major{primarily} focus our comparison on more practical FU methods. Halimi et al. \cite{halimi2022federated} propose Project Gradient Ascent (PGA) to erase client influence by constraining local weight updates within an L2-norm ball around a reference model. While PGA does not require historical updates or auxiliary data, it introduces other constraints. Clients must be stateful, storing their latest model updates, and full participation is mandatory, as unlearning relies on a reference model obtained by removing the last target client’s update from the current global model. Additionally, directly applying gradient ascent on forget data risks gradient explosion, as the loss function often lacks an upper bound, leading to divergence and poor generalization. To counteract this, Halimi et al. employ gradient clipping and L2-norm constraints \cite{halimi2022federated}. However, gradient-ascent techniques require careful and extensive hyperparameter tuning, such as selecting the appropriate L2-norm radius or gradient clipping threshold. \major{\tool{} eliminates the risk of uncontrolled gradient growth while maintaining the standard local training procedure. This is achieved by rescaling the client’s local update (pseudo-gradient) after training, rather than modifying the training dynamics.} \major{MoDe \cite{zhao2023federated} introduces a multi-round unlearning protocol alternating between degradation and memory-guidance phases. This requires additional per-round coordination among clients with respect to FedAvg, as well as more per-round computational and communication requirements.
In contrast, \tool{} does not require any change in the regular FedAvg training round.
FedAU \cite{gu2024unlearning} trains a local auxiliary module on forget data with randomized labels, and forms the unlearned model as a linear combination of the original model and the auxiliary module. This mechanism requires clients to train an auxiliary module during the regular FedAvg  rounds, hence introducing a variation over the base protocol.} Recently, Khalil et al. \cite{khalil2025not} proposed NoT, a novel method aiming to achieve unlearning without impractical assumptions by implementing server-side unlearning. Specifically, NoT inverts the sign of the first layer’s global model parameters upon receiving an unlearning request. While this approach does not require explicit client-side unlearning, it fails to remove a target client's contribution selectively. In fact, NoT produces the same unlearned model regardless of which client exits the federation. Our experimental evaluation (see Figure \ref{fig:radar_intro} and Section \ref{sec:experimental_results}) shows that NoT requires a longer recovery period and fails to match the forgetting performance of full model retraining.

In contrast to previous methods, \tool{} offers several key advantages. It does not rely on historical updates or auxiliary data, does not require full client participation, and operates in a stateless manner. Moreover, \tool{} is task-agnostic, meaning it does not need modifications based on the specific learning task. It maintains the same computational cost as standard FedAvg rounds and requires minimal tuning, as demonstrated in our experimental results. \major{Furthermore, \tool{} supports sample unlearning and can handle multiple unlearning requests simultaneously.} These properties make \tool{} a practical and effective solution for federated unlearning, inherently aligned with the design principles of FL.


%% file: sections/background.tex
\section{Background}\label{sec:background}

\subsection{Federated Learning}
FedAvg \cite{mcmahan2017communication} is a widely used baseline in FL. It adopts a client-server framework in which a central server manages the global model parameters. Training proceeds in synchronous rounds, where a randomly selected subset of client devices is invited to participate. During each round, the activated clients fine-tune the global model using their local data over a fixed number of epochs. The server then collects the locally computed updates from the clients, and incorporates them into the global model $w$, before distributing the updated global parameters again.

In FedAvg, clients typically send model updates to the server. These updates are calculated as the difference between the global weights, received at the beginning of the round, and the locally trained weights. This can be expressed mathematically as:

\begin{equation}
\label{eq:fedavg_delta1}
w_{t+1} = \frac{1}{|S_t|}\sum_{i \in S_t} w_t^i = w_t + \frac{1}{|S_t|}\sum_{i \in S_t} (w_t^i - w_t)
\end{equation}

where $S_t$ represents the set of clients selected for round $t$, $w$ represents the weight of the global model and $w^i$ are the weights computed locally at client $i$. For clarity, in Equation \ref{eq:fedavg_delta1}, model updates are shown without being scaled by the number of local data samples, but this simplification does not affect the subsequent analysis. By defining client updates as $\Delta_t^i := w_t^i - w_t$ and their aggregated form as:

\begin{equation} 
\Delta_t := \frac{1}{|S_t|} \sum_{i \in S_t} \Delta_t^i,
\end{equation} 

and by introducing a (global) learning rate $\eta_s$, the FedAvg's update rule can be rewritten as:

\begin{equation} 
\label{eq:pseudo_g} 
w_{t+1} = w_t + \eta_s \Delta_t = w_t -\eta_s (- \Delta_t) 
\end{equation}

This shows that the server’s update rule in FedAvg is equivalent to applying SGD to the \emph{pseudo-gradient} $- \Delta_t$ with $\eta_s=1$. This perspective reveals that FedAvg is a specific instance of the broader FedOpt framework \cite{reddi2020adaptive}, which can use various server-side optimizers. \major{Building on this view, Reddi et al. \cite{reddi2020adaptive} formally treat $-\Delta_t$ as a stochastic gradient in their FedOpt framework and prove convergence guarantees for a wide range of server optimizers (e.g., SGD, Adam, Yogi) under standard smoothness and bounded-variance assumptions. This establishes that the aggregated client update has the same mathematical role as a true gradient in the server’s optimization step, despite the server never accessing raw data. }

\subsection{Federated Unlearning}
An FU method aims to remove specific learned information from the global model $w$ while preserving the \emph{good} knowledge acquired on data that should not be forgotten. FU objectives are classified based on the information the algorithm is expected to forget: sample unlearning, class unlearning \cite{wang2023classdiscriminative}, feature unlearning \cite{gu2025ferrari}, and client unlearning. Our work mainly focuses on client unlearning.

During a training round $t$, a client $u$ may withdraw previous contributions from the global model $w$. This is achieved through an unlearning procedure $\mathcal{U}(w_t,D_u)$, which can be executed by different entities within the FL framework, such as the server, the client to be forgotten, or the remaining clients \cite{liu2024survey}. An unlearning algorithm should produce a novel version of the original global model $w_t^{\overline{u}}$, called the unlearned model, that effectively excludes the influence of the forget data $D_u$. The method is effective if $w_t^{\overline{u}}$ exhibits performances approximately indistinguishable from the retrained model $w_t^r$, a model trained as if client $u$ had never joined the federation.

The efficiency of FU is usually measured by the number of rounds needed to obtain the generalization performance of $w_t^r$. Effectiveness in forgetting the client data $D_u$ is assessed by comparing the performance of $w_t^{\overline{u}}$ and $w_t^r$ on $D_u$. Common metrics for verifying the success of unlearning include loss, accuracy, and susceptibility to membership inference attacks (MIAs) \cite{shokri2017membership,hu2022membership,rahman2018membership}. MIAs test whether certain samples were used at training time, \carlo{by providing a measure of the unlearning effectiveness in removing traces of such samples.}

\subsection{Terminology} 
In this subsection, we define key terminology that will be used throughout the paper.  

\begin{itemize}
    \item \textbf{target client:} The client that requests unlearning, also referred to as the \textit{Target Client} or client $u$. The objective of an unlearning algorithm is to remove the influence of client $u$ data (i.e., \major{forget} data) from the global model.

    \item \textbf{Original Model:} The global model trained with the participation of both the target client and all other clients.  

    \item \textbf{Retrained Model:} The global model trained without the target client from the beginning of the FL process. This serves as the gold-standard baseline for evaluating the effectiveness of unlearning methods.  

    \item \textbf{Unlearned Model:} The original model after the unlearning procedure (for example, after applying \tool{}). Since unlearning may impact model performance, a recovery phase is often necessary to restore its generalization capabilities to match the retrained model. As detailed in Section \ref{sec:experimental_results}, an optimal FU algorithm should minimize the discrepancy with the retrained model in terms of forgetting metrics (e.g., accuracy on client $u$'s data, susceptibility to MIAs) while achieving equal or superior generalization to the retrained model in the fewest possible FL rounds.
\end{itemize}

%% file: sections/method.tex
\section{\tool{}: Federated Unlearning via Negated Pseudo-Gradients}\label{sec:approach}


\subsection{Preliminaries}
Before presenting our original method, we provide the necessary preliminaries and notation to fully and more easily understand our proposal. During a given round $t$, a subset of the participating clients may request the removal of their contributions. We define $S_t^-$ as the subset of clients requesting unlearning, and $S_t^+$ as the remaining clients. \carlo{Additionally, let $n$ denote the total number of samples held by participating clients \carlo{during} the unlearning round, such that:}

\begin{equation} 
n := \sum_{i \in S_t^+}|D_i| + \sum_{j \in S_t^-}|D_j|
\end{equation}

We define two aggregated model updates: $\Delta_t^+$ , representing the aggregated updates contributed by the clients in $S_t^+$, and $\Delta_t^-$, representing the aggregated updates contributed by the clients in $S_t^-$, as follows:

 \begin{equation} 
    \label{eq:fedavg_delta_plus} 
    \Delta_t^+ := \frac{1}{n} \sum_{i \in S_t^+} \Delta_t^i = \frac{1}{n} \sum_{i \in S_t^+} |D_i|  (w_t^i - w_t)
    \end{equation}
    
    \begin{equation} 
    \label{eq:fedavg_delta_minus} 
    \Delta_t^- := \frac{1}{n} \sum_{j \in S_t^-} \Delta_t^j = \frac{1}{n} \sum_{j \in S_t^-} |D_j|  (w_t^j - w_t)
\end{equation}

We also define the forget data (or target data) as the union of the local dataset held by clients that request unlearning, expressed as $D_u:= \bigcup_{j \in S_t^-} D_j$.


\begin{figure*}[!t]
\centering
\includegraphics[width=0.99\textwidth]{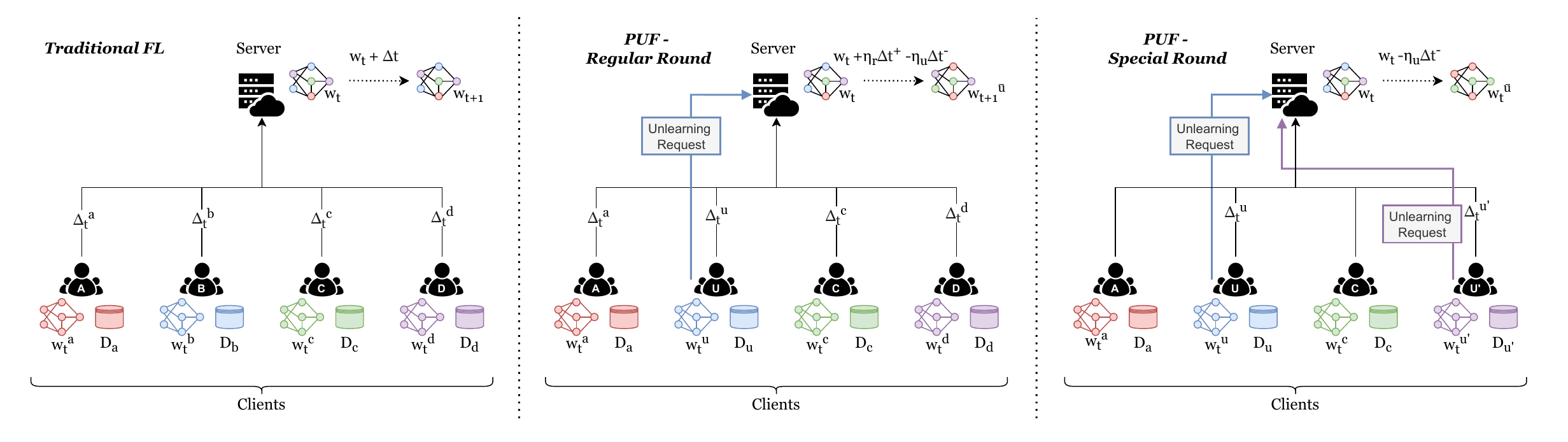}
\caption{Overview of \tool{} operating modes. In \tool{}-Regular Round, target clients provide their model updates together with other clients. In \tool{}-Special Round, only target clients share their model updates.}
\label{fig:method}
\end{figure*}

\subsection{Federated Unlearning via Negated Pseudo-Gradients}
In FedAvg \cite{mcmahan2017communication}, clients receive the weights of the current global model and send back model updates. As discussed in Section \ref{sec:background}, these updates can be interpreted as pseudo-gradients, also enabling the use of server-side optimizers \cite{reddi2020adaptive}. \major{This interpretation is not merely conceptual: in the FedOpt framework of Reddi et al.~\cite{reddi2020adaptive}, the server update in FedAvg is mathematically equivalent to applying a gradient-based optimizer to the pseudo-gradient $-\Delta_t$, with formal convergence guarantees under standard assumptions.} 
We leverage this interpretation for unlearning in \tool{}. Specifically, \tool{} flips the sign of the model update \major{(pseudo-gradient)} computed by the target client(s), applying it in the opposite direction of the local optimization. Figure \ref{fig:method} provides an illustrative overview \major{of} our method.

This approach emulates the effect of gradient ascent-based unlearning techniques by pushing the global model parameters away from regions where the loss function on the \major{forget} data is minimized. A key technical advantage of \tool{} over directly applying gradient ascent, which maximizes the loss function rather than minimizing it on the \major{forget} data, is that it avoids the risk of gradient explosion. Since the loss function generally lacks an upper bound, gradient-ascent techniques can lead to uncontrolled gradient growth, preventing convergence and severely degrading the model’s generalization ability. To address this, regularization methods such as weight projection (e.g., L2-norm ball) and gradient clipping are often employed to constrain parameter updates. However, these techniques require careful and often extensive hyperparameter tuning, including selecting an appropriate L2-norm radius or gradient clipping threshold. 

In contrast, our method eliminates the risk of uncontrolled gradient growth by preserving the standard local training procedure on clients while negating and rescaling the resulting update (pseudo-gradient) after local training. \tool{} \carlo{supports} two alternative \carlo{operating} modes:
\begin{itemize}
    \item \textbf{Regular Round with Modified Aggregation (\tool{}-Regular).} In this mode, unlearning is integrated into a regular training round, where both unlearning and remaining clients participate. When one or more clients opt to leave the federation and be forgotten, the round proceeds as usual, except that the aggregated model \carlo{updates} from target clients are negated. The sign flip can be applied either locally at clients, before transmitting back the updates, or during the server-side aggregation. The updates from other clients remain unchanged. The resulting aggregation can be formalized as follows:

    \begin{equation} 
    \label{eq:fedavg_delta3} 
   \Delta_t = \eta_r \Delta_t^+ - \eta_u \Delta_t^-,
    \end{equation}
    with $\eta_r$ and $\eta_u$ being global learning rates. In particular, $\eta_u$ scales the aggregated update from target clients. Considering SGD as a server-side optimizer and $\eta_s=1$ (see Eq. \ref{eq:pseudo_g}), the resulting unlearning model is generated as:
    
    \begin{equation} 
    \label{eq:fedavg_delta3} 
    w_{\bar{u}} = w_t + \Delta_t
    \end{equation}
    Regular FL training resumes from $w_{\bar{u}}$. 
    
    \item \textbf{Special Unlearning Round with Only target clients (\tool{}-Special).} In this mode, a dedicated unlearning round is conducted before resuming regular training. Only the target clients participate, retrieving the latest global model, and performing regular training. Assuming SGD as the server-side optimizer with $\eta_s = 1$, the aggregated update is negated and applied to the global model as follows:

\begin{equation} \label{eq:aggregation_single} w_{\bar{u}} = w_t - \eta_u \Delta_t^-, \end{equation}

where $\eta_u$ is the scaling factor for the aggregated update from target clients. Regular FL training resumes from $w_{\bar{u}}$.
\end{itemize}

\input{algorithms/puf_algorithm}

\noindent Algorithm \ref{alg:puf} reports a unified framework for both \tool{}-Special and \tool{}-Regular. For readability, at line 14, we assume SGD with $\eta_s$=1.0 as the server-side optimizer. 
\major{It is important to emphasize that \tool{}-Regular and \tool{}-Special are \emph{alternative} unlearning modes, and are not meant to be used jointly or sequentially. For any given unlearning request, only one of the two routines should be applied. We included both variants to support different operational contexts: \tool{}-Special aligns with existing works (e.g., \cite{halimi2022federated}), which isolate unlearning into dedicated rounds, while \tool{}-Regular allows for seamless integration into standard FL workflows.}

\subsection{\major{Discussion}}

The effectiveness of \tool{} depends on the magnitude of the unlearning updates, which must be carefully controlled to prevent excessive model drift. As we discuss in Section \ref{sec:experimental_results}, the tuning of the scaling factor $\eta_u$  is crucial for balancing unlearning effectiveness while mitigating degradation in the model's generalization performance.
\tool{} efficiently removes contributions from target clients while remaining fully compatible with the traditional FedAvg. After the unlearning round, the global model may require a few additional rounds to recover its generalization performance, as observed in all approximate FU mechanisms \cite{romandini2024federated}. It is worth noting that in both \carlo{operating} modes, clients can transmit model updates to the server, as done in traditional FedAvg, without requiring \carlo{any} modifications \carlo{or additional overhead}. 
\tool{} is designed to ensure the unlearning process is task-agnostic and seamless for federated participants, supporting multiple concurrent unlearning requests. 

\begin{table*}[t!]
\scriptsize
\centering
\adjustbox{max width=0.99\textwidth}{
    \begin{tabular}{llcclcccc}
        \toprule
        Model & Dataset & Clients & \major{Unlearning Case} & Data Heterogeneity & Pre-trained & $E$ & Task & \# target clients\\
        \midrule
        ResNet-18 & CIFAR-10 & 10 & Client & LDA ($\alpha$=0.3) & $\times$ & 1 &  $\mathcal{C}$ & Up to one\\
        ResNet-18 & CIFAR-100 & 10 & Client & LDA ($\alpha$=0.1) or IID & $\times$ & 1 or 10 & $\mathcal{C}$ & Up to two\\
        ResNet-18 & CIFAR-100 & 10 & Sample & LDA ($\alpha$=0.1) or IID & $\times$ & 1 or 10 & $\mathcal{C}$ & Up to one\\
        MiT-B0 & CIFAR-100 & 10 & Client & LDA ($\alpha$=0.1) & \checkmark & 1 & $\mathcal{C}$ & Up to two \\
        ResUNET & ProstateMRI & \major{6} & Client & Real-world feature skew & $\times$ & 1 & $\mathcal{S}$ & Up to one\\
        \bottomrule
    \end{tabular}
}
\caption{Settings of reported results. $\mathcal{C}$=classification task, $\mathcal{S}$=segmentation task. $E$=local epochs during FedAvg. \textit{\# target clients} reports if the experiments considers a single client or multiple clients to unlearn.}
\label{tab:outline_experiments}
\end{table*}


\smallskip
\noindent \major{\textbf{Sample Unlearning.}}
\major{Our work mainly focuses on client unlearning. However, \tool{} can be readily extended to sample unlearning scenarios. In this setting, during the unlearning round, the clients that request to be forgotten use only the subset of their local dataset that must be removed when computing the model update.}

\smallskip
\noindent \major{\textbf{Limitation.}}
\major{As in prior FU methods such as \cite{federaser, halimi2022federated, zhao2023federated, gu2024unlearning}, \tool{} assumes that the target clients participate one final time in the unlearning phase to generate the scaled and negated pseudo-gradient update. This design reflects a deliberate trade-off: by requiring a last participation, we completely avoid storing historical model updates or client states, preserving the ephemeral nature of FL. Moreover, in cases where the client invokes its right to be forgotten for privacy reasons, our approach allows the removal of its contribution without retaining any of its past updates, thus aligning with data minimization principles.}



%% file: algorithms/puf_algorithm.tex
\begin{algorithm}[t]
\caption{\small\tool{} Algorithm. Note that $S_t^+ = \emptyset$ when PUF-Special is used for the unlearning phase, while $S_t^+ \neq \emptyset$ and $S_t^- \neq \emptyset$ for PUF-Regular. $S_t^- = \emptyset$ for regular training rounds. }
\begin{algorithmic}[1]
\Require{Global model weights $w$, local epochs $E$, global learning rate $\eta_s$, unlearning rate $\eta_u$, local learning rate $\eta$, batch size $B$}
\State Initialize $w_0$
\For{each round $t = 0, 1, 2, \dots$}
    \If{\major{$t$ \textbf{is} unlearning round}} 
    \State $S_t^+ \leftarrow$ (random set of remaining clients)
    \State 
    \Comment{$S_t^+ = \emptyset$} if PUF-Special 
    \State $S_t^- \leftarrow$ (set of target clients)
    \Else \Comment{\major{Standard FedAvg Round}}
    \State \major{$S_t^+ \leftarrow$ (random set of clients)}
    \State \major{$S_t^- = \emptyset$}
    \EndIf
    \For{each client $i \in S_t^+$ \textbf{simultaneously}}
        \State $\Delta_{t}^i \leftarrow$ \textbf{ClientOpt}$(i, w_t)$
    \EndFor
    \For{each client $j \in S_t^-$ \textbf{simultaneously}}
        \State $\Delta_{t}^j \leftarrow$ \textbf{ClientOpt}$(j, w_t)$
    \EndFor
    \State $\Delta_t^+ \leftarrow \frac{1}{n} \sum_{i \in S_t^+} \Delta_t^i$
    \State $\Delta_t^- \leftarrow \frac{1}{n} \sum_{j \in S_t^-} \Delta_t^j$
    \State $\Delta_t \leftarrow \eta_r \Delta_t^+ - \eta_u \Delta_t^-$
    \State $w_{t+1} \leftarrow w_t + \Delta_t$
\EndFor

\Procedure{\textbf{ClientOpt$(k, w_t)$}}{}
    \State $w \leftarrow w_t$
    \State $\mathcal{B} \leftarrow$ (split $\mathcal{D}_k$ into batches of size $B$)
    \For{each local epoch $e$ from 1 to $E$}
        \For{each batch $b \in \mathcal{B}$}
            \State $w \leftarrow w - \eta\nabla \ell(w;b) $
        \EndFor
    \EndFor
    \State $\Delta_t^k \leftarrow w - w_t$ 
    \State \textbf{return} $\Delta_t^k$ to server
\EndProcedure
\end{algorithmic}
\label{alg:puf}
\end{algorithm}

%% file: sections/results.tex
\input{figures/data_distribution}

\section{Experimental Results}
\label{sec:experimental_results}

\subsection{Datasets, Models, and Learning Setting}
We conducted a comprehensive set of experiments on image classification tasks and image segmentation tasks. For image classification, we used federated versions of the CIFAR-10 and CIFAR-100 datasets \cite{Krizhevsky09learningmultiple}, which consist of 60,000 32×32 color images. CIFAR-10 comprises 10 classes, while CIFAR-100 includes 100 classes. The datasets were partitioned to simulate 10 clients, ensuring no overlapping samples among them. The experiments were performed under both Identically and Independently Distributed (IID) and non-IID settings across clients. To create non-IID data, we applied label-skew partitioning based on a distribution determined by Latent Dirichlet Allocation (LDA) \cite{hsu2019measuring}, using concentration parameters of $\alpha=0.3$ for CIFAR-10 and $\alpha=0.1$ for CIFAR-100. Figure \ref{fig:cifar100_distrib} provides a visual depiction of the label distribution among clients. For image classification, we employed a standard ResNet-18 \cite{he2016deep} and a visual transformer, namely the MiT-B0 \cite{xie2021segformer}. Before performing unlearning, we train ResNet-18 from scratch for 200 FL rounds while we fine-tune MiT-B0 for 50 FL rounds, starting from a pre-trained checkpoint. If not differently indicated, clients run one local epoch ($E=1$) for standard FedAvg. 

For image segmentation, we conducted a set of experiments using the ProstateMRI federated dataset  \cite{liu2020ms} that comprises prostate T2-weighted MRI scans (with segmentation masks) sourced from six data providers, each one treated as a client. Due to variations in imaging protocols across sites, real-world feature heterogeneity naturally arises among clients. The dataset includes 384x384 color images alongside their corresponding segmentation masks.

\major{We also design a set of experiments for sample unlearning, using the same configurations and hyperparameters as in the client unlearning setup. When evaluating sample unlearning, 50\% of the samples in the local dataset of each client that requests to be forgotten are removed. The forget subset is obtained through uniform random selection from the client's local dataset. The remaining samples (retain data) are then used in the recovery phase for all considered methods, meaning that the target clients contribute to the recovery process only with their retain data.}

Table \ref{tab:outline_experiments} outlines the settings of our experiments. For each setting, we performed a variable \major{number} of experiments, considering one or more clients to forget.
In all experiments, we consider SGD as a server-side optimizer with $\eta_s$=1.0, and we set $\eta_r$=1.0 for PUF-Regular. If not differently indicated, we set $\eta_u$=2.0 for \tool{}-Special and $\eta_u$=20.0 for \tool{}-Regular (Section \ref{sec:hp_tuning} discusses the tuning of our methods). \major{Further} details about per-setting hyper-parameters are reported in the following.

\smallskip
\noindent\textbf{ResNet-18 on CIFAR-10/CIFAR-100.} We used a standard ResNet-18 \cite{he2016deep} and employed Group Normalization layers, similar to other works with similar settings (e.g., \cite{kim2022multi}). We used SGD as a local optimizer with a learning rate set to 0.1, with a round-wise exponential decay of 0.998, 1 or 10 local epochs ($E=1$ or $E=10$), local batch size of 32. We pre-processed the training images with random crop, horizontal flip and normalization layers. During unlearning routines, we only apply normalization. 

\smallskip
\noindent\textbf{MiT-B0 on CIFAR-100.} We used a visual transformer, i.e., MiT-B0 \cite{xie2021segformer}, with approximately 3.6M parameters, initialized from a pre-trained model checkpoint trained on ImageNet-1k (69.27\% accuracy on test data). We adapted the one-layer classification head to this task, initializing such a layer from scratch. We employed the AdamW optimizer with a client learning rate of 3e-4, with a round-wise exponential decay of 0.998, 1 local epoch, local batch size of 32, and weight decay regularization of 1e-3. The images are resized to a resolution of 224x224. 

\smallskip
\noindent\textbf{Res-UNet on ProstateMRI.} We utilized a vanilla Res-UNet architecture \cite{yu2017volumetric}, similar to the one used in \cite{liu2020ms}, with approximately 7.6M parameters. In line with \cite{zhou2023fedfa}, we trained the network using a combination of standard cross-entropy and Dice loss \cite{milletari2016v}, and we conducted regular training for 500 rounds, before applying unlearning. We used Adam  \cite{kingma2014adam} as local optimizer with a client learning rate of 1e-4, a local batch size of 16, and a local weight decay of 1e-4.

\subsection{Baselines}

We compared \tool{} against \major{five} state-of-the-art baseline methods, which are \major{FedEraser \cite{federaser},} PGA \cite{halimi2022federated}, \major{FedAU \cite{gu2024unlearning}, MoDe \cite{zhao2023federated}} and NoT \cite{khalil2025not}. \major{Our comparisons primarily focus on FU methods that avoid impractical assumptions, such as the need for historical information. Nevertheless, we also include FedEraser \cite{federaser}, despite its requirement to store past client models, as it serves both as a widely adopted reference baseline in the literature and as a means to illustrate the overhead incurred by methods that rely on this requirement.} \major{Hyper-parameter tuning of baseline methods can be found in Appendix A (Supplementary Material).}

\major{As PGA \cite{halimi2022federated} and NoT \cite{khalil2025not} were originally evaluated with full client participation, both during the training of the original model and in the recovery phase, we adopted the same setting to ensure a fair comparison.}

In some experiments, we included a \textit{Natural} baseline, which bypasses any unlearning procedure before the recovery phase. This allows us to assess whether the influence of the target client would naturally vanish over time without intervention.


\subsection{Metrics}
We evaluated the proposed methods along two key dimensions: \major{their \emph{efficacy} in achieving unlearning and their \emph{efficiency} in recovering model utility}.  
\major{For unlearning efficacy}, the primary goal is to minimize the discrepancy between the metrics of the approximate unlearning method and those of the gold-standard retrained model. In this context, neither higher nor lower absolute metric values are inherently preferable; instead, the objective is to reduce the absolute difference with respect to retraining from scratch. We denote this absolute difference with the prefix $\Delta$ (e.g., $\Delta$ Test Accuracy represents the absolute difference in Test Accuracy between the unlearned model and the retrained model).  
\major{For unlearning efficiency, we measure the cumulative \emph{computational} (FLOPs), \emph{communication} (bytes), and \emph{storage} (bytes) overhead required to perform unlearning and to recover model utility. In addition, we report the \emph{relative improvement} over retraining from scratch, where higher gains indicate better performance.}

\smallskip
\noindent\textbf{Test Accuracy ($\Delta$ Test Acc.).} We evaluated the test accuracy at two critical stages: immediately after the unlearning routine (before the recovery phase) and after the recovery phase concludes. We used the standard test set provided with the dataset, which is IID and contains data that have never been accessed or observed by any client during training.

\smallskip
\noindent\textbf{Forget Accuracy ($\Delta$ Forget Acc.).} We evaluated the unlearned model's accuracy on client $u$'s train data and computed the absolute difference relative to the retrained model's performance. This metric provides insight into how closely the unlearning process approximates the desired outcome achieved through retraining.

\smallskip
\noindent\textbf{MIAs on Forget Data ($\Delta$ MIA).} The MIA aims to 
infer whether specific samples were part of the training data for the attacked model. Its success rate quantifies the fraction of target data, defined as ($D_u:= \bigcup_{j \in S_t^-} D_j$), correctly identified as belonging to the training set. A lower MIA success rate indicates reduced information leakage about $D_u$ from the attacked model. To implement MIAs, we employed two established approaches: (i) a confidence-based MIA predictor \cite{song2021systematic}, and (ii) a loss-based MIA predictor \cite{yeom2018privacy}.  

\begin{enumerate}
    \item Confidence-based MIA predictor \cite{song2021systematic}: It consists of a training phase using a balanced dataset of seen and unseen data. The retain dataset serves as the seen data, while the standard test dataset is used as the unseen data. An MIA predictor is then trained to classify whether the target model's output corresponds to seen or unseen data based on prediction confidence.
    \item Loss-based MIA predictor \cite{yeom2018privacy}: It assumes the attacker can access the target model’s average training loss. Samples are classified as members of the training set if their loss falls below this average; otherwise, they are labeled as non-members. For our federated implementation, we assume access to the global mean training loss.
\end{enumerate}

\smallskip
\noindent\major{\textbf{Communication, Computational, and Storage Costs.} We compute cumulative communication, computational, and storage costs for each federated unlearning method, considering both the unlearning phase and the recovery phase.
In the unlearning phase, we quantify these costs according to the original design of each algorithm, under consistent assumptions between methods.
The recovery phase corresponds to standard FedAvg rounds involving only retained clients; all methods incur the same per-round costs during recovery, differing only in the number of recovery rounds required to restore model utility.
For retraining baselines, the reported costs correspond to training the model from scratch with the remaining clients, i.e., the total FedAvg rounds performed after removing the target clients. We provide a detailed explanation of the overhead calculations for each method in Appendix B (Supplementary Material).}


\input{tables/results_after_review}

\subsection{Main Results}

\smallskip
\noindent\major{\textbf{Client Unlearning.}}
\major{Table \ref{table:full_results} reports the performance of \tool{}’s best configurations (hyper-parameter tuning in Sec.~\ref{sec:hp_tuning}) and other baselines across multiple datasets (CIFAR-100 and CIFAR-10), data distributions (IID and Non-IID), and model architectures (ResNet-18, MiT-B0) in the single-client unlearning scenario. Results are averaged over ten independent runs, each with a different target client.
Across all settings, \tool{} attains the lowest or near-lowest $\Delta$ values in forget metrics (e.g., Forget Accuracy), meaning it most closely replicates the retraining gold standard, thereby achieving highly effective forgetting.}

\major{In \textbf{CIFAR-100 (IID, ResNet-18)}, PUF-Special and PUF-Regular reduce $\Delta$ Forget Accuracy to below 5\%, outperforming MoDe (11.1\%), NoT (12.6\%), and FedAU (16.8\%). While FedAU achieves the highest efficiency overall (communication and computation $\approx$60$\times$ lower than retraining), this comes at the cost of poor forgetting performance. Conversely, FedEraser delivers almost perfect forgetting ($\Delta\approx$1.5\%), but with efficiency close to retraining in communication and computation, and drastically higher storage needs. PUF variants strike a balance, offering sub-5\% $\Delta$ while maintaining $31\text{--}48\times$ communication and computation gains.}

\major{In \textbf{CIFAR-100 (Non-IID, ResNet-18, $E$=1)}, PUF-Regular achieves the best $\Delta$ Forget Accuracy (2.0\%), ahead of FedAU (4.5\%), MoDe (8.4\%), and NoT (9.9\%), while preserving $25\times$ efficiency gains over retraining. PUF-Special slightly sacrifices forgetting (3.3\% $\Delta$) for even higher efficiency ($\approx40\times$). PGA achieves higher efficiency ($\approx28\times$) but worse forgetting (4.3\% $\Delta$ Forget Accuracy).}

\major{In \textbf{CIFAR-100 (Non-IID, ResNet-18, $E$=10)}, PUF-Regular exhibits the lowest $\Delta$ Forget Accuracy (1.7\%) with $33\times$ efficiency gains, while PUF-Special offers the highest efficiency ($\approx56\times$) and competitive forgetting (2.4\%). PGA is also efficient ($\approx44\times$) but less accurate in forgetting (3.1\%).}

\major{In \textbf{CIFAR-10 (Non-IID, ResNet-18)}, PUF-Special reaches the best forgetting performance ($\Delta$ Forget Accuracy $<$1\%) with the highest efficiency ($\approx49\times$), outperforming all baselines in both dimensions.}

\major{Finally, in \textbf{CIFAR-100 (Non-IID, MiT-B0)}, PUF-Special attains near-perfect forgetting ($\Delta$=0.4\%) with solid efficiency ($\approx4.5\times$), while PGA is the most efficient ($\approx7\times$) but exhibits notably weaker forgetting ($\Delta$=4.4\%).}
\major{Overall, PUF variants consistently achieve an effective trade-off between efficacy and efficiency.}


\input{figures/recovery_phase}

\smallskip
\noindent\textbf{Performance Consistency During Recovery.}
Figure \ref{fig:recovery_detail} illustrates the evolution of Test Accuracy and Forget Accuracy throughout the recovery phase, beginning at recovery round 0, which corresponds to the application of the unlearning routine (either PGA, \major{MoDe, FedAU,} NoT, or our proposed solution, \tool{}). We do not report the evolution of the considered MIAs, as they closely follow the trends of Forget Accuracy.
The figure presents results for three different settings, as indicated in the subcaptions. For reference, we also include values for both the Original Model and the Retrained Model. We emphasize the following key aspects for the full and easy understanding of the reported charts:
\begin{enumerate}
    \item The recovery phase concludes when the unlearned model surpasses the Test Accuracy of the retrained model for the first time.
    \item A more effective forgetting method should minimize the gap in Forget Accuracy at the end of the recovery phase.
    \item A better forgetting method should also reduce the discrepancy with the retrained model's Test Accuracy during the recovery phase, ensuring higher usability of the unlearned model, which serves as the new FL global model for inference during these rounds.
\end{enumerate}
Thus, Figure \ref{fig:recovery_detail} provides valuable insights into the dynamics of these metrics during the recovery phase. The analysis highlights that PUF produces a superior unlearned model, significantly narrowing the gap with the retrained model earlier in the recovery phase compared to the other baselines. The figure clearly shows that the other baselines fail to achieve accurate forgetting by the end of the recovery phase, as evidenced by their larger gaps in Forget Accuracy. \major{All the methods show an accuracy drop after unlearning. This is an inherent consequence of strategies that operate directly on the global model, as the removal of a client's contributions inevitably results in a temporary degradation of overall performance \cite{romandini2024federated}.}

\input{tables/sample_unlearning}

\input{tables/results_multiple_after_review}

\smallskip
\noindent\major{\textbf{Sample Unlearning.} Table~\ref{tab:efficacy_not} reports the results for sample-level unlearning when 50\% of each client’s dataset is designated as forget data. Across both IID and non-IID CIFAR-100 settings, \tool{} consistently delivers the smallest $\Delta$ Forget Accuracy, confirming its superior forgetting effectiveness over all baselines. In the IID setting, \tool{}-Special achieves a $\Delta$ Forget Accuracy of only $6.0\%$, outperforming FedAU ($7.2\%$) and substantially better than NoT ($13.9\%$). Similar trends are observed in the auxiliary membership inference metrics (MIA$_{\text{Song}}$ and MIA$_{\text{Yeom}}$), where \tool{}-Special reduces the $\Delta$ to $4.6\%$, the lowest among all methods. \tool{}-Regular matches this forgetting effectiveness closely ($\Delta$ Forget Accuracy $6.1\%$), still outperforming FedAU and clearly surpassing NoT. In the non-IID scenario, \tool{}-Special maintains a $\Delta$ Forget Accuracy of only $2.2\%$, significantly ahead of FedAU ($6.9\%$) and NoT ($8.8\%$). Also \tool{}-Regular, at $4.3\%$, outperforms both baselines by a clear margin. These results indicate that PUF’s update-scaling strategy remains effective under higher data heterogeneity, where other methods experience substantial degradation in forgetting performance. 
Concerning efficiency, both PUF variants achieve large reductions in cumulative communication and computation costs compared to Retrain, up to $200.5\times$ and $189.3\times$ for \tool{}-Special in the non-IID setting. While FedAU attains the highest cost reduction ($222.7\times$ communication, $199.3\times$ computation), this comes at the expense of weaker forgetting (e.g., $\Delta$ Forget Accuracy $6.9\%$ in non-IID vs. $2.2\%$ for \tool{}-Special). NoT, conversely, exhibits both lower efficiency (e.g., $27.9\times$ communication gain in non-IID) and weaker forgetting effectiveness. Overall, the results confirm that \tool{} offers a favorable balance between forgetting accuracy and efficiency in sample unlearning, matching or exceeding the best efficiency reductions while preserving retrain-level forgetting performance.
}

\input{tables/segmentation_table}

\smallskip
\noindent\major{\textbf{Multiple Unlearning.} 
Table \ref{table:multiple_unlearning_after_reviews} presents the performance on federated CIFAR-100 (non-IID) using a ResNet-18 or an MiT-B0 architecture, where a pair of randomly chosen clients was designated as target clients (averaged over five independent experiments). We do not include PGA as a baseline because it is not directly extensible to unlearning multiple tasks. This limitation arises from its reliance on a reference model, which is obtained by removing the last stored model update of the target client. As a result, the mutual dependency of unlearning updates prevents its straightforward adaptation to multiple unlearning requests. Similarly, we do not include MoDe, since its memory-guidance phase is not easily adaptable to this scenario. }

\major{For ResNet-18, PUF-Regular achieves the most faithful replication of retraining, with a Forget Accuracy gap of just 2.9\%, accompanied by equally low deviations in both MIA metrics. PUF-Special follows closely, with only a modest increase in the forgetting gap (4.0\%) but delivering the best efficiency gains of all methods, cutting communication and computation costs by more than a factor of 58. By contrast, FedAU maintains good forgetting performance (5.7\% gap) and offers notable savings (18$\times$), but still falls short of PUF in both accuracy preservation and privacy metrics. NoT shows clear signs of incomplete forgetting, with a gap exceeding 9\%, which is also reflected in higher MIA scores.}

\major{The pattern is consistent for MiT-B0. PUF-Regular exhibits the smallest gap to retraining (2.3\%), outperforming FedAU (4.6\%) by nearly half and dramatically surpassing NoT, which exhibits a large 15.6\% gap. PUF-Special maintains solid forgetting (6.3\%) while achieving higher efficiency than FedAU, but slightly less than NoT, which remains the most communication-efficient method here (15.6$\times$) at the cost of very weak forgetting.}


\input{figures/hp_tuning_figures}

\smallskip
\noindent\textbf{Task Agnosticism.} As highlighted in Sections \ref{sec:related} and \ref{sec:approach}, the design of PUF ensures that the unlearning phase remains task-agnostic, as both the unlearning and remaining clients follow the standard training procedure without any modifications to their local learning routines. Table \ref{table:prostate_unet_after_review} reports the results for the ProstateMRI experiments, which focus on a segmentation task rather than classification. \major{FedAU is not applicable to segmentation tasks like Res-UNet, as it requires a separate classification head, which such architectures lack. MoDe’s memory-guidance phase is also not clearly adaptable to non-classification tasks, with no instructions in the original paper.} PUF outperforms all baselines in terms of both recovery efficiency (i.e., shorter recovery phase) and forgetting efficacy. As discussed in Section \ref{sec:hp_tuning}, PUF can be further tuned for improved forgetting performance by increasing the scaling factor $\eta_u$, albeit at the cost of a longer recovery phase. Notably, NoT does not offer this tuning flexibility.

\subsection{Hyper-parameter Tuning of \tool{}}
\label{sec:hp_tuning}
Figure \ref{fig:hp_tuning_puf} compares the unlearning performance ($\Delta$ Forget Accuracy on the Y-axis, recovery rounds on the X-axis) of various configurations of PUF-Special and PUF-Regular with the best configurations of NoT and PGA, and the Natural baseline. The analysis also includes configurations where PUF employs more local epochs (denoted as $E_u$ in Figure \ref{fig:hp_tuning_puf}; when omitted, $E_u=1$). Points closer to the origin of the axes indicate better performance. For this analysis, we focused exclusively on the $\Delta$ Forget Accuracy metric, as the MIAs metrics align closely with this metric (as emerges from tabular results). We highlight in red the configurations we found to be the best and used them in the Experimental Results section.

Notably, increasing local epochs ($E_u > 1$) for \tool{} reduces $\Delta$ Forget Accuracy but requires more recovery rounds compared to $E_u=1$ with the same $\eta_u$. Interestingly, increasing the scaling factor $\eta_u$ while keeping $E_u=1$ achieves similar improvements without additional local computational overhead for target clients, making it more practical in resource-constrained settings. Both PUF-Regular and PUF-Special achieve high effectiveness by simply tuning the scaling factor ($\eta_u$) while maintaining $E_u=1$.  As shown in Figure \ref{fig:hp_tuning_puf}, \tool{} demonstrates low sensitivity to the $\eta_u$ hyper-parameter, resulting in a cluster of well-performing configurations.

In contrast, we observe that other baseline methods require multiple hyper-parameter to tune. For example, PGA necessitates adjusting the local learning rate, local epochs, an early stopping threshold, and gradient-clipping threshold, while also assuming full client participation and stateful clients. \major{MoDE involves tuning the number of memory-guidance and degradation rounds, separate learning rates for each, and a degradation parameter controlling momentum. FedAU needs to tune the weight to assign to the auxiliary module when merging it with the original model, as well as the number of epochs to train the module.} On the other hand, NoT does not require specific hyperparameter tuning but fails to achieve significant forgetting, as shown in Table \ref{table:full_results}. 
This highlights \tool{}'s trade-off between ease of integration, ease of hyperparameter optimization, and unlearning performance compared to other baselines.

\input{figures/hp_tuning_segmentation}

Figure \ref{fig:hp_segm} shows the hyper-parameter tuning of PUF for the ProstateRMI dataset, on segmentation data and real-world feature heterogeneity. Also in this setting, PUF exhibits low sensitivity to its main hyper-parameter.


%% file: figures/data_distribution.tex
\begin{figure*}[t!]
\centering

\begin{subfigure}[t]{0.32\textwidth} %
    \centering   
    \includegraphics[width=0.95\linewidth]{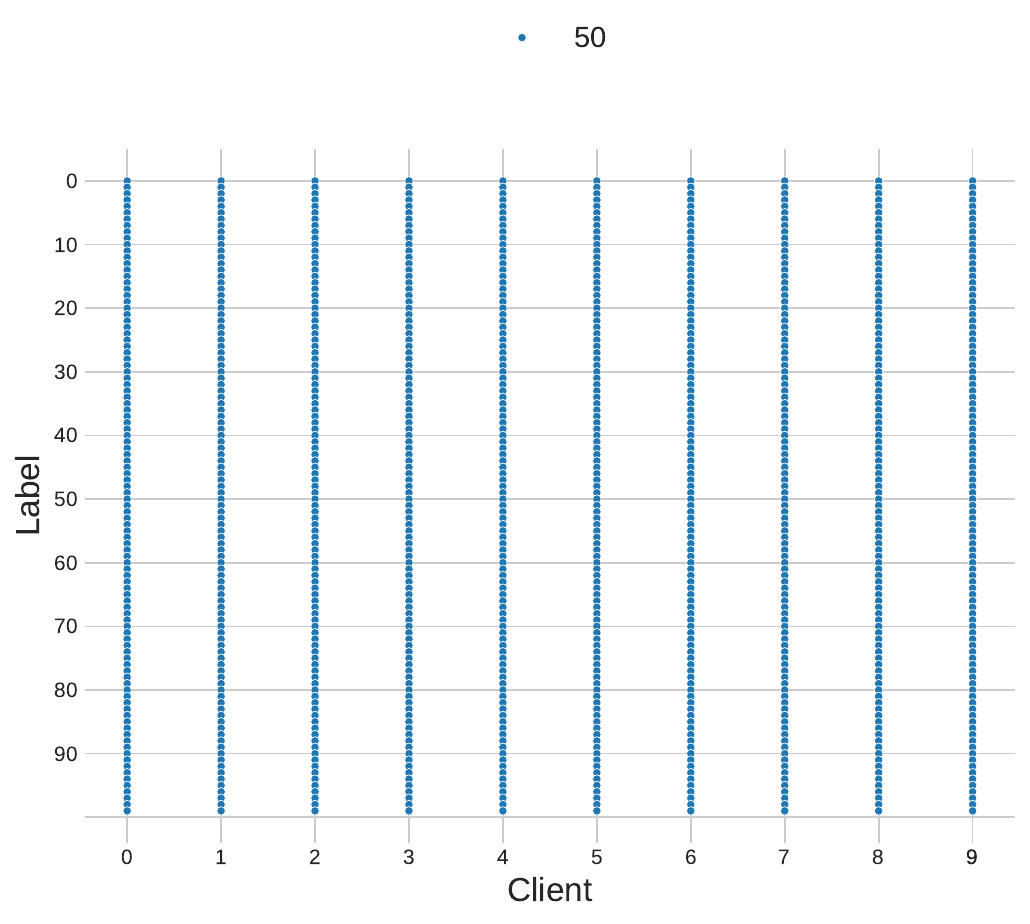}
    \caption{CIFAR-100 (IID).}
    \label{}
\end{subfigure}
\hfill
\begin{subfigure}[t]{0.32\textwidth} %
    \centering   
    \includegraphics[width=0.95\linewidth]{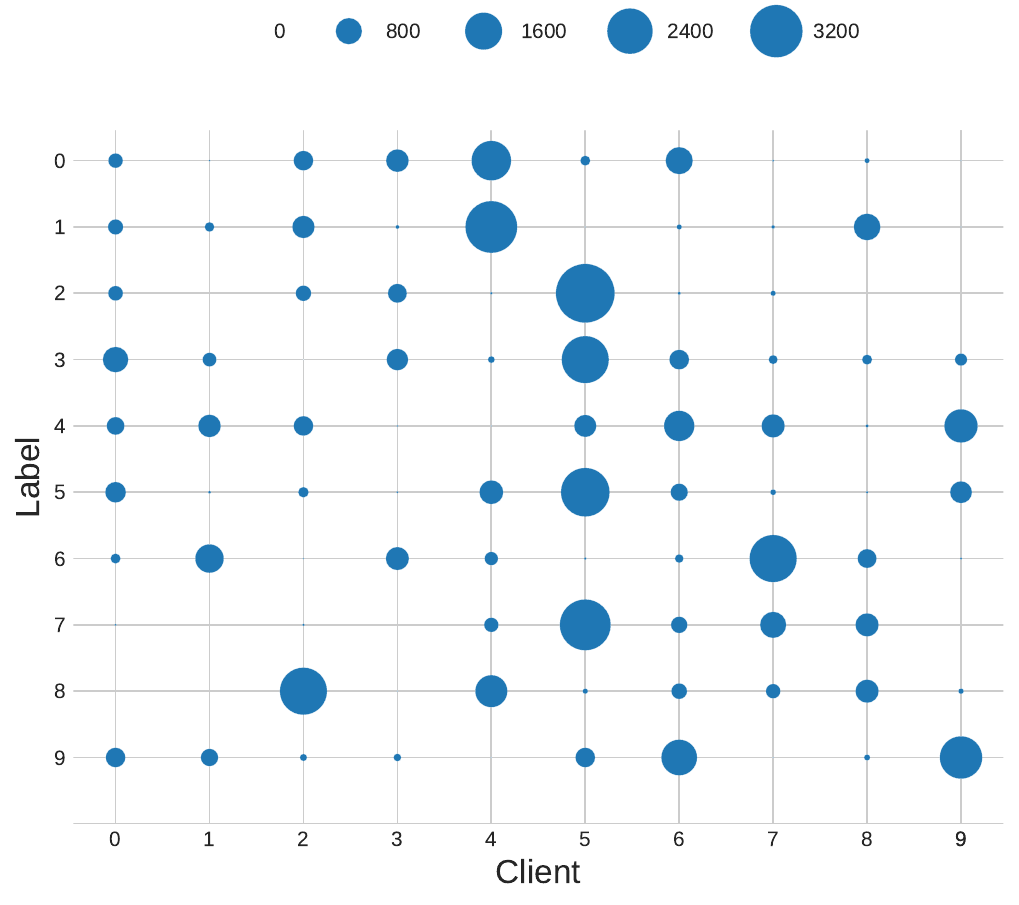}
    \caption{CIFAR-10 (Non-IID, $\alpha=0.3$).}
    \label{}
\end{subfigure}
\hfill
\begin{subfigure}[t]{0.32\textwidth} %
    \centering   
    \includegraphics[width=0.95\linewidth]{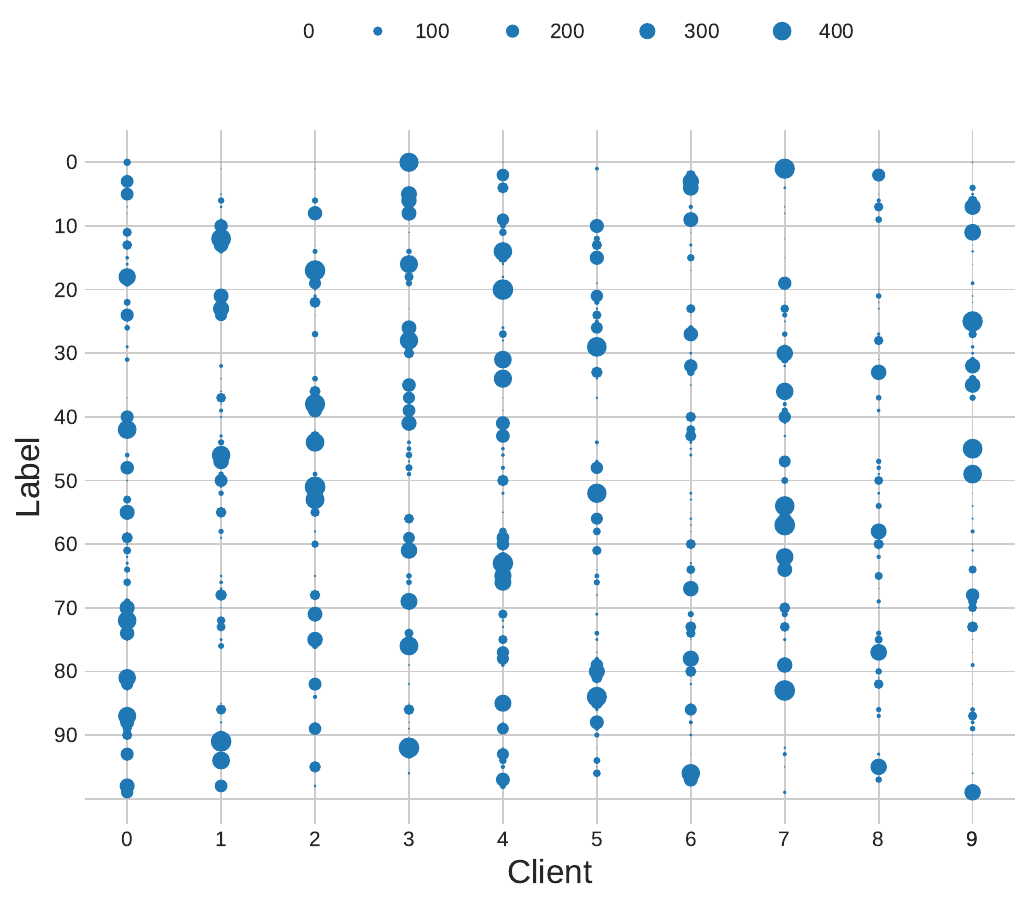}
    \caption{CIFAR-100 (Non-IID, $\alpha=0.1$).}
    \label{}
\end{subfigure}

\caption{Label distribution across clients (0-9) for  CIFAR-100 (IID), CIFAR-10 (Non-IID, $\alpha=0.3$), and CIFAR-100 (Non-IID$, \alpha=0.1$).}
\label{fig:cifar100_distrib}
\end{figure*}

%% file: tables/results_after_review.tex
\begin{table*}[!t]
\small
\centering
\adjustbox{max width=0.98\textwidth}{
\begin{tabular}{llccccccc}
\toprule
\multirow{4}{*}{\makecell[l]{\textbf{Setting}}} & \multirow{4}{*}{\makecell[l]{\textbf{Method}}} & \multicolumn{4}{c}{\textbf{Efficacy}} & \multicolumn{3}{c}{\textbf{Efficiency}} \\
\cmidrule(lr){3-6} \cmidrule(lr){7-9}
& & \textbf{Test Acc.} & \textbf{Forget Acc. ($\Delta$ $\downarrow$)} & \textbf{MIA$_{[Song]}$ ($\Delta$ $\downarrow$)} & \textbf{MIA$_{[Yeom]}$ ($\Delta$ $\downarrow$)} & \textbf{Communication} & \textbf{Computation} & \textbf{Storage} \\
& & & & & & Bytes ($\times$ $\uparrow$) & FLOPs ($\times$ $\uparrow$) & Bytes ($\times$ $\uparrow$) \\
\midrule
\multirow{9}{*}{\makecell[l]{\textbf{CIFAR-100},\\ IID,\\ ResNet-18,\\ $E$=1}} 
& Original & 59.9$_{\pm 0.0}$ & 79.8$_{\pm 0.7}$ & 78.2$_{\pm 0.6}$ & 71.4$_{\pm 0.6}$ & --- & --- & --- \\
& Retrain & 58.3$_{\pm 0.5}$ & 58.0$_{\pm 0.7}$ & 55.6$_{\pm 0.7}$ & 48.8$_{\pm 0.6}$ & 1.62e$^{11}$ (1.0×) & 1.35e$^{15}$ (1.0×) & 4.49e$^{07}$ (1.0×) \\
&    FedEraser & 58.4$_{\pm 0.5}$ & \underline{59.5 (1.5$_{\pm 0.7}$)} & \underline{57.6 (2.0$_{\pm 0.5}$)} & \underline{49.9 (1.1$_{\pm 1.1}$)} & 1.62e$^{11}$ (1.0×) & 6.75e$^{14}$ (2.0×) & 8.98e$^{10}$ (0.0005×) \\
& PGA & 59.1$_{\pm 0.6}$ & 72.8 (14.8$_{\pm 1.0}$) & 71.8 (16.1$_{\pm 1.8}$) & 63.0 (14.1$_{\pm 1.6}$) & 9.86e$^{09}$ (16.4×) & 8.54e$^{13}$ (15.8×) & 4.94e$^{08}$ (0.0909×) \\
&    MoDe & 58.7$_{\pm 0.5}$ & 69.0 (11.1$_{\pm 0.7}$) & 70.2 (14.6$_{\pm 2.0}$) & 62.4 (13.5$_{\pm 0.8}$) & 1.75e$^{10}$ (9.3×) & 1.46e$^{14}$ (9.3×) & 8.98e$^{07}$ (0.5×) \\
&    FedAU & 59.0$_{\pm 0.5}$ & 74.7 (16.8$_{\pm 1.0}$) & 76.3 (20.6$_{\pm 2.1}$) & 61.6 (12.8$_{\pm 2.4}$) & \textbf{2.67e$^{09}$ (60.8×)} & \textbf{2.23e$^{13}$ (60.6×)} & 4.49e$^{07}$ (1.0×) \\
& NoT & 58.9$_{\pm 0.4}$ & 70.5 (12.6$_{\pm 0.8}$) & 69.3 (13.7$_{\pm 1.9}$) & 63.3 (14.4$_{\pm 0.8}$) & 5.25e$^{09}$ (30.9×) & 4.39e$^{13}$ (30.8×) & 4.49e$^{07}$ (1.0×) \\
& PUF-Special & 58.8$_{\pm 0.3}$ & 62.8 (4.9$_{\pm 1.1}$) & 61.5 (5.9$_{\pm 1.4}$) & 55.8 (7.0$_{\pm 1.1}$) & 3.40e$^{09}$ (47.6×) & 2.84e$^{13}$ (47.5×) & 4.49e$^{07}$ (1.0×) \\
& PUF-Regular & 58.8$_{\pm 0.4}$ & \textbf{62.4 (4.4$_{\pm 1.0}$)} & \textbf{60.8 (5.2$_{\pm 0.9}$)} & \textbf{55.5 (6.7$_{\pm 1.1}$)} & 5.10e$^{09}$ (31.8×) & 4.26e$^{13}$ (31.7×) & 4.49e$^{07}$ (1.0×) \\
\midrule
\multirow{9}{*}{\makecell[l]{\textbf{CIFAR-100},\\ Non-IID,\\ ResNet-18,\\ $E$=1}} 
& Original & 53.8$_{\pm 0.0}$ & 62.9$_{\pm 6.9}$ & 75.4$_{\pm 7.0}$ & 61.0$_{\pm 7.8}$ & --- & --- & --- \\
& Retrain & 51.0$_{\pm 1.4}$ & 33.5$_{\pm 4.5}$ & 44.0$_{\pm 5.5}$ & 32.0$_{\pm 5.2}$ & 1.62e$^{11}$ (1.0×) & 1.35e$^{15}$ (1.0×) & 4.49e$^{07}$ (1.0×) \\
&    FedEraser & 51.2$_{\pm 0.5}$ & \underline{35.7 (2.2$_{\pm 1.6}$)} & \underline{49.1 (3.4$_{\pm 2.0}$)} & \underline{47.4 (3.4$_{\pm 1.5}$)} & 1.62e$^{11}$ (1.0×) & 6.75e$^{14}$ (2.0×) & 8.98e$^{10}$ (0.0005×) \\

& PGA & 51.4$_{\pm 0.6}$ & 37.4 (4.3$_{\pm 4.1}$) & 49.2 (6.4$_{\pm 4.9}$) & 36.6 (5.3$_{\pm 5.3}$) & 
5.74e$^{09}$ (28.2×) & 5.10e$^{13}$ (26.5×) & 4.94e$^{08}$ (0.09×) \\
&    MoDe & 50.9$_{\pm 1.1}$ & 41.9 (8.4$_{\pm 3.8}$) & 55.0 (10.6$_{\pm 4.0}$) & 42.3 (10.3$_{\pm 3.8}$) & 2.19e$^{10}$ (7.4×) & 1.83e$^{14}$ (7.4×) & 8.98e$^{07}$ (0.5×) \\
&   FedAU & 51.2$_{\pm 1.2}$ & 38.4 (4.5$_{\pm 2.2}$) & 51.7 (6.7$_{\pm 2.5}$) & 38.3 (5.3$_{\pm 1.9}$) & 1.29e$^{10}$ (12.5×) & 1.08e$^{14}$ (12.5×) & 4.49e$^{07}$ (1.0×) \\
& NoT & 51.3$_{\pm 0.2}$ & 43.3 (9.9$_{\pm 4.4}$) & 55.7 (11.7$_{\pm 5.7}$) & 44.3 (12.3$_{\pm 5.5}$) & 1.03e$^{10}$ (15.7×) & 8.64e$^{13}$ (15.6×) & 4.49e$^{07}$ (1.0×) \\
& PUF-Special & 52.3$_{\pm 0.4}$ & 36.8 (3.3$_{\pm 2.2}$) & \textbf{47.3 (3.4$_{\pm 2.8}$)} & \textbf{35.9 (3.9$_{\pm 2.6}$)} & \textbf{4.01e$^{09}$ (40.4×)} & \textbf{3.56e$^{13}$ (38.0×)} & 4.49e$^{07}$ (1.0×) \\
& PUF-Regular & 52.2$_{\pm 0.6}$ & \textbf{35.4 (2.0$_{\pm 2.0}$)} & 48.8 (4.8$_{\pm 2.4}$) & 36.4 (4.4$_{\pm 2.3}$) & 6.44e$^{09}$ (25.1×) & 5.59e$^{13}$ (24.1×) & 4.49e$^{07}$ (1.0×) \\
\midrule
\multirow{9}{*}{\makecell[l]{\textbf{CIFAR-100},\\ Non-IID,\\ ResNet-18,\\ $E$=10}} 
& Original & 50.3$_{\pm 0.0}$ & 60.8$_{\pm 6.6}$ & 73.4$_{\pm 6.6}$ & 61.3$_{\pm 7.5}$ & --- & --- & --- \\
& Retrain & 48.0$_{\pm 0.8}$ & 31.6$_{\pm 4.5}$ & 42.1$_{\pm 5.2}$ & 31.2$_{\pm 4.9}$ & 1.62e$^{11}$ (1.0×) & 1.35e$^{16}$ (1.0×) & 4.49e$^{07}$ (1.0×) \\
& PGA & 46.3$_{\pm 1.9}$ & 34.7 (3.1$_{\pm 3.9}$) & 45.3 (3.2$_{\pm 4.0}$) & 33.5 (2.3$_{\pm 3.7}$) & 3.01e$^{14}$ (44.9×) & 3.35e$^{13}$ (40.4×) & 4.94e$^{08}$ (0.09×) \\
&    MoDe & 46.9$_{\pm 2.3}$ & 35.3 (3.9$_{\pm 3.6}$) & 45.4 (4.7$_{\pm 4.1}$) & 34.0 (3.6$_{\pm 3.3}$) & 2.10e$^{10}$ (7.7×) & 1.76e$^{15}$ (7.7×) & 8.98e$^{07}$ (0.5×) \\
&    FedAU & 46.8$_{\pm 2.4}$ & 36.8 (4.5$_{\pm 3.7}$) & 47.1 (6.1$_{\pm 3.6}$) & 34.7 (2.8$_{\pm 3.5}$) & 1.13e$^{10}$ (14.3×) & 9.45e$^{14}$ (14.3×) & 4.49e$^{07}$ (1.0×) \\
& NoT & 46.7$_{\pm 2.2}$ & 39.7 (8.1$_{\pm 3.6}$) & 48.9 (6.8$_{\pm 4.4}$) & 37.2 (6.0$_{\pm 3.9}$) & 6.22e$^{09}$ (26.0×) & 5.20e$^{14}$ (26.0×) & 4.49e$^{07}$ (1.0×) \\
& PUF-Special & 47.8$_{\pm 1.5}$ & 34.0 (2.4$_{\pm 2.9}$) & \textbf{43.7 (1.6$_{\pm 3.2}$)} & \textbf{32.4 (1.2$_{\pm 3.0}$)} & \textbf{2.90e$^{09}$ (55.8×)} & \textbf{2.67e$^{14}$ (50.7×)} & 4.49e$^{07}$ (1.0×) \\
& PUF-Regular & 47.6$_{\pm 1.6}$ & \textbf{33.3 (1.7$_{\pm 2.8}$)} & 44.5 (2.4$_{\pm 2.9}$) & 32.7 (1.5$_{\pm 2.7}$) & 4.85e$^{09}$ (33.4×) & 4.44e$^{14}$ (30.4×) & 4.49e$^{07}$ (1.0×) \\
\midrule
\multirow{9}{*}{\makecell[l]{\textbf{CIFAR-10},\\ Non-IID,\\ ResNet-18,\\ $E$=1}} 
& Original & 83.7$_{\pm 0.0}$ & 88.6$_{\pm 3.9}$ & 84.4$_{\pm 5.4}$ & 81.4$_{\pm 6.3}$ & --- & --- & --- \\
& Retrain & 83.5$_{\pm 1.6}$ & 81.1$_{\pm 8.0}$ & 73.1$_{\pm 13.7}$ & 72.5$_{\pm 10.8}$ & 1.62e$^{11}$ (1.0×) & 1.35e$^{15}$ (1.0×) & 4.49e$^{07}$ (1.0×) \\
& PGA & 84.0$_{\pm 1.0}$ & 84.3 (3.2$_{\pm 4.9}$) & 78.5 (5.4$_{\pm 7.7}$) & 77.0 (4.5$_{\pm 6.0}$) & 9.38e$^{09}$ (17.3×) & 8.14e$^{13}$ (16.6×) & 4.94e$^{08}$ (0.09×) \\
&    MoDe & 83.5$_{\pm 1.6}$ & 82.3 (2.0$_{\pm 1.7}$) & 72.6 (2.6$_{\pm 1.8}$) & 71.8 (1.5$_{\pm 1.0}$) & 2.40e$^{10}$ (6.7×) & 2.01e$^{14}$ (6.7×) & 8.98e$^{07}$ (0.5×) \\
&    FedAU & 83.8$_{\pm 1.4}$ & 85.2 (4.0$_{\pm 2.6}$) & 76.9 (4.3$_{\pm 3.7}$) & 73.2 (2.9$_{\pm 2.9}$) & 5.17e$^{09}$ (31.3×) & 4.32e$^{13}$ (31.2×) & 4.49e$^{07}$ (1.0×) \\
& NoT & 83.8$_{\pm 1.2}$ & 84.3 (3.1$_{\pm 3.0}$) & 78.0 (4.9$_{\pm 4.5}$) & 75.3 (2.9$_{\pm 3.4}$) & 1.37e$^{10}$ (11.9×) & 1.14e$^{14}$ (11.8×) & 4.49e$^{07}$ (1.0×) \\
& PUF-Special & 83.9$_{\pm 1.2}$ & \textbf{82.0 (0.9$_{\pm 1.4}$)} & \textbf{72.0 (1.4$_{\pm 1.7}$)} & \textbf{71.1 (2.1$_{\pm 2.0}$)} & \textbf{3.30e$^{09}$ (49.1×)} & \textbf{2.79e$^{13}$ (48.5×)} & 4.49e$^{07}$ (1.0×) \\
& PUF-Regular & 83.8$_{\pm 1.2}$ & 82.4 (1.3$_{\pm 1.6}$) & 72.8 (2.2$_{\pm 2.0}$) & 71.5 (2.4$_{\pm 1.8}$) & 5.31e$^{09}$ (30.5×) & 4.49e$^{13}$ (30.1×) & 4.49e$^{07}$ (1.0×) \\
\midrule
\multirow{9}{*}{\makecell[l]{\textbf{CIFAR-100},\\ Non-IID,\\ MiT-B0,\\ $E$=1}} 
& Original & 75.0$_{\pm 0.0}$ & 84.3$_{\pm 5.6}$ & 77.0$_{\pm 8.6}$ & 74.0$_{\pm 6.4}$ & --- & --- & --- \\
& Retrain & 73.3$_{\pm 0.8}$ & 57.8$_{\pm 5.3}$ & 48.6$_{\pm 5.0}$ & 44.6$_{\pm 3.9}$ & 1.19e$^{10}$ (1.0×) & 3.82e$^{15}$ (1.0×) & 1.33e$^{07}$ (1.0×) \\
& PGA & 73.6$_{\pm 0.4}$ & 62.2 (4.4$_{\pm 3.3}$) & 53.6 (5.0$_{\pm 3.7}$) & 50.5 (6.9$_{\pm 3.2}$) & \textbf{1.68e$^{09}$ (7.1×)} & \textbf{5.70e$^{14}$ (6.7×)} & 1.46e$^{08}$ (0.09×) \\
&    MoDe & 73.4$_{\pm 0.8}$ & 62.9 (5.2$_{\pm 2.2}$) & 54.7 (6.1$_{\pm 3.1}$) & 50.3 (5.9$_{\pm 2.9}$) & 5.55e$^{09}$ (2.1×) & 1.78e$^{15}$ (2.2×) & 2.66e$^{07}$ (0.5×) \\
&    FedAU & 73.6$_{\pm 0.7}$ & 63.7 (5.9$_{\pm 1.9}$) & 55.2 (5.9$_{\pm 2.3}$) & 50.2 (5.6$_{\pm 1.3}$) & 3.54e$^{09}$ (3.4×) & 1.13e$^{15}$ (3.4×) & 1.33e$^{07}$ (1.0×) \\
& NoT & 73.4$_{\pm 0.7}$ & 67.9 (10.1$_{\pm 4.1}$) & 59.4 (10.1$_{\pm 6.6}$) & 54.0 (9.4$_{\pm 5.4}$) & 3.70e$^{09}$ (3.2×) & 1.19e$^{15}$ (3.2×) & 1.33e$^{07}$ (1.0×) \\
& PUF-Special & 73.7$_{\pm 0.5}$ & \textbf{58.2 (0.4$_{\pm 1.5}$)} & \textbf{47.2 (1.4$_{\pm 1.7}$)} & \textbf{45.1 (0.5$_{\pm 1.4}$)} & 2.65e$^{09}$ (4.5×) & 8.42e$^{14}$ (4.5×) & 1.33e$^{07}$ (1.0×) \\
& PUF-Regular & 73.6$_{\pm 0.6}$ & 59.3 (1.5$_{\pm 1.7}$) & 48.4 (2.7$_{\pm 1.8}$) & 45.8 (1.2$_{\pm 1.3}$) & 3.92e$^{09}$ (3.0×) & 1.26e$^{15}$ (3.0×) & 1.33e$^{07}$ (1.0×) \\

\bottomrule
\end{tabular}
}
\caption{\major{Performance of PUF and other baselines across different settings for client unlearning. Baseline results are expressed as \textit{mean metric value (mean $\Delta$ $\pm$ standard deviation)}, with $\Delta$ in parentheses representing the average absolute difference from the Retrain baseline. Lower $\Delta$ corresponds to better unlearning performances. For efficiency metrics, higher $\times$ (reduction over Retrain baseline) are better. The best-performing method is shown in bold. When FedEraser \cite{federaser} ranks first, it is underlined, as its efficiency limitations make it impractical in realistic scenarios.}}
\label{table:full_results}
\end{table*}
\captionsetup{labelfont={color=black}}

%% file: figures/recovery_phase.tex

\begin{figure*}[t!]
\centering

\begin{subfigure}[t]{0.48\textwidth} 
    \centering
    \includegraphics[width=0.49\linewidth]{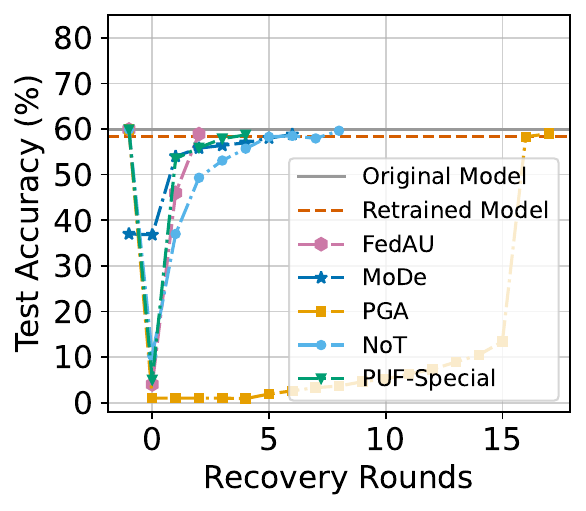}
    \includegraphics[width=0.49\linewidth]{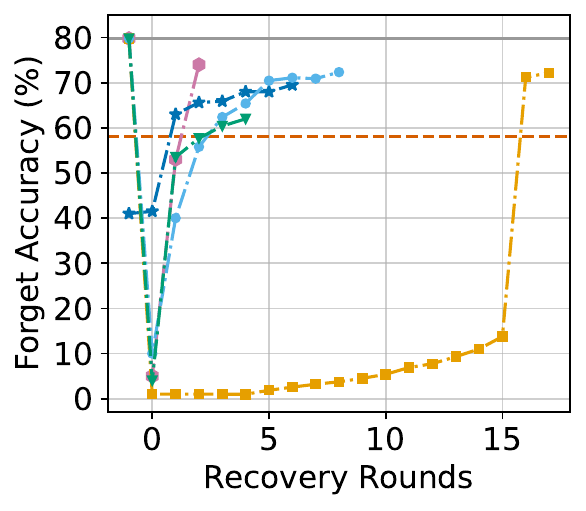}
    \caption{CIFAR-100, IID, ResNet-18.}
\end{subfigure}
\hfill
\begin{subfigure}[t]{0.48\textwidth}
    \centering
    \includegraphics[width=0.49\linewidth]{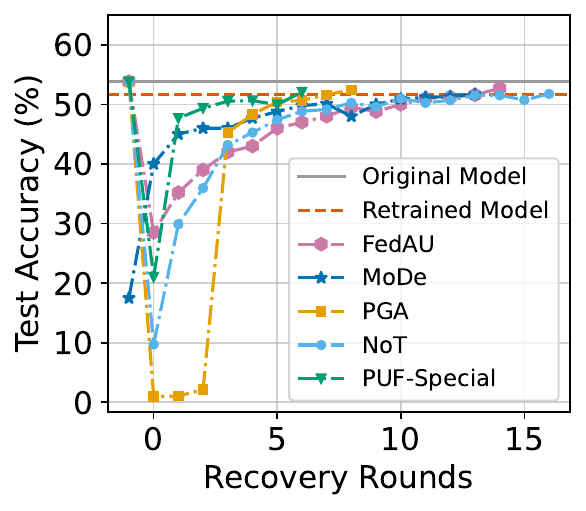}
    \includegraphics[width=0.49\linewidth]{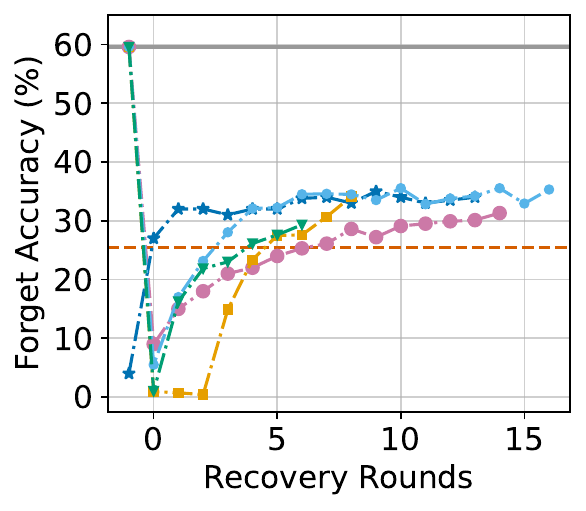}
    \caption{CIFAR-100, Non-IID, ResNet-18.}
\end{subfigure}

\caption{Evolution of generalization ability (test accuracy) and forgetting effectiveness (forget accuracy) during the recovery phase across three different settings. Each pair of images presents Test Accuracy (\textbf{Left}) and Forget Accuracy (\textbf{Right}) for a representative client in a specific setting indicated in subcaption as a triple \textit{dataset, data distribution, model architecture}. For our method, the charts only report the performance of PUF-Special for better visualization. For MoDe we do not show the multi-round unlearning phase for clarity. }
\label{fig:recovery_detail}
\end{figure*}
\captionsetup{labelfont={color=black}}

%% file: tables/sample_unlearning.tex
\begin{table*}[!t]
\small
\centering
\adjustbox{max width=0.98\textwidth}{
\begin{tabular}{llccccccc}
\toprule
\multirow{4}{*}{\makecell[l]{\textbf{Setting}}} & \multirow{4}{*}{\makecell[l]{\textbf{Method}}} & \multicolumn{4}{c}{\textbf{Efficacy}} & \multicolumn{3}{c}{\textbf{Efficiency}} \\
\cmidrule(lr){3-6} \cmidrule(lr){7-9}
& & \textbf{Test Acc.} & \textbf{Forget Acc. ($\Delta$ $\downarrow$)} & \textbf{MIA$_{[Song]}$ ($\Delta$ $\downarrow$)} & \textbf{MIA$_{[Yeom]}$ ($\Delta$ $\downarrow$)} & \textbf{Communication} & \textbf{Computation} & \textbf{Storage} \\
& & & & & & Bytes ($\times$ $\uparrow$) & FLOPs ($\times$ $\uparrow$) & Bytes ($\times$ $\uparrow$) \\
\midrule

\multirow{5}{*}{\makecell[l]{\textbf{CIFAR-100},\\ IID,\\ ResNet-18,\\ $E$=1}} 
&   Retrain & 59.4$_{\pm 0.4}$ & 58.5$_{\pm 0.5}$ & 56.0$_{\pm 0.6}$ & 49.2$_{\pm 0.7}$ & 1.80e$^{11}$ (1.0×) & 1.42e$^{15}$ (1.0×) & 4.49e$^{07}$ (1.0×) \\
&    FedAU & 59.8$_{\pm 0.3}$ & 65.7 (7.2$_{\pm 1.1}$) & 61.8 (5.8$_{\pm 0.9}$) & 54.9 (5.7$_{\pm 0.9}$) & \textbf{8.08e$^{08}$ (222.7×)} & \textbf{7.12e$^{12}$ (199.3×)} & 4.49e$^{07}$ (1.0×) \\
&   NoT & 59.7$_{\pm 0.5}$ & 72.4 (13.9$_{\pm 0.7}$) & 68.1 (12.1$_{\pm 1.0}$) & 60.7 (11.5$_{\pm 1.0}$) & 5.65e$^{09}$ (31.8×) & 4.99e$^{13}$ (28.5×) & 4.49e$^{07}$ (1.0×) \\
&   \tool{}-Special & 60.0$_{\pm 0.8}$ & \textbf{64.5 (6.0$_{\pm 1.1}$)} & \textbf{60.6 (4.6$_{\pm 0.9}$)} & \textbf{53.8 (4.6$_{\pm 0.9}$)} & 1.43e$^{09}$ (125.8×) & 1.22e$^{13}$ (116.4×) & 4.49e$^{07}$ (1.0×) \\
&   \tool{}-Regular & 60.1$_{\pm 0.8}$ & 64.6 (6.1$_{\pm 1.2}$) & 60.7 (4.7$_{\pm 0.9}$) & 53.9 (4.7$_{\pm 0.9}$) & 2.51e$^{09}$ (71.6×) & 2.18e$^{13}$ (65.3×) & 4.49e$^{07}$ (1.0×) \\

\midrule
\multirow{5}{*}{\makecell[l]{\textbf{CIFAR-100},\\ Non-IID,\\ ResNet-18,\\ $E$=1}} 
&   Retrain & 49.0$_{\pm 0.6}$ & 40.3$_{\pm 2.4}$ & 38.9$_{\pm 2.0}$ & 34.1$_{\pm 1.8}$ & 1.80e$^{11}$ (1.0×) & 1.42e$^{15}$ (1.0×) & 4.49e$^{07}$ (1.0×) \\
&    FedAU & 50.0$_{\pm 0.2}$ & 47.2 (6.9$_{\pm 1.2}$) & 45.1 (6.2$_{\pm 1.1}$) & 40.2 (6.1$_{\pm 1.1}$) & \textbf{8.08e$^{08}$ (222.7×)} & \textbf{7.12e$^{12}$ (199.3×)} & 4.49e$^{07}$ (1.0×) \\
&   NoT & 49.6$_{\pm 0.8}$ & 49.0 (8.8$_{\pm 1.4}$) & 47.0 (8.1$_{\pm 1.3}$) & 41.8 (7.7$_{\pm 1.2}$) & 6.46e$^{09}$ (27.9×) & 5.70e$^{13}$ (24.9×) & 4.49e$^{07}$ (1.0×) \\
&   \tool{}-Special & 50.1$_{\pm 1.0}$ & \textbf{42.5 (2.2$_{\pm 1.2}$)} & \textbf{40.9 (2.0$_{\pm 1.1}$)} & \textbf{36.3 (2.2$_{\pm 1.1}$)} & 8.98e$^{08}$ (200.5×) & 7.50e$^{12}$ (189.3×) & 4.49e$^{07}$ (1.0×) \\
&   \tool{}-Regular & 50.2$_{\pm 1.0}$ & 41.3 (4.3$_{\pm 1.3}$) & 40.1 (4.0$_{\pm 1.2}$) & 35.2 (4.1$_{\pm 1.2}$) & 1.71e$^{09}$ (105.5×) & 1.46e$^{13}$ (97.1×) & 4.49e$^{07}$ (1.0×) \\

\bottomrule
\end{tabular}
}
\caption{\major{Performance of PUF and other baselines across different settings for sample unlearning when half of the dataset is the forget data. For efficacy metrics, baseline results are expressed as \textit{mean metric value (mean $\Delta$ $\pm$ standard deviation)}, with $\Delta$ in parentheses representing the average absolute difference from the Retrain baseline. Lower $\Delta$ corresponds to better unlearning performances. For efficiency metrics, higher $\times$ (reduction over Retrain baseline) are better. The best-performing method is shown in bold.}}
\label{tab:efficacy_not}
\end{table*}
 \captionsetup{labelfont={color=black}} 

%% file: tables/results_multiple_after_review.tex
\begin{table*}[!t]
\small
\centering
\adjustbox{max width=0.98\textwidth}{
\begin{tabular}{llccccccc}
\toprule
\multirow{4}{*}{\makecell[l]{\textbf{Setting}}} & \multirow{4}{*}{\makecell[l]{\textbf{Method}}} & \multicolumn{4}{c}{\textbf{Efficacy}} & \multicolumn{3}{c}{\textbf{Efficiency}} \\
\cmidrule(lr){3-6} \cmidrule(lr){7-9}
& & \textbf{Test Acc.} & \textbf{Forget Acc. ($\Delta$ $\downarrow$)} & \textbf{MIA$_{[Song]}$ ($\Delta$ $\downarrow$)} & \textbf{MIA$_{[Yeom]}$ ($\Delta$ $\downarrow$)} & \textbf{Communication} & \textbf{Computation} & \textbf{Storage} \\
& & & & & & Bytes ($\times$ $\uparrow$) & FLOPs ($\times$ $\uparrow$) & Bytes ($\times$ $\uparrow$) \\
\midrule

\multirow{6}{*}{\makecell[l]{\textbf{Multiple Unlearning},\\ CIFAR-100, Non-IID,\\ ResNet-18, $E$=1}}
& Original     & 53.8$_{\pm 0.0}$ & 63.1$_{\pm 6.8}$ & 76.0$_{\pm 7.6}$ & 61.1$_{\pm 8.3}$ & --- & --- & --- \\
& Retrain      & 47.3$_{\pm 2.1}$ & 30.1$_{\pm 3.3}$ & 39.7$_{\pm 3.9}$ & 27.9$_{\pm 3.1}$ & 1.44e$^{11}$ (1.0×) & 1.20e$^{15}$ (1.0×) & 4.49e$^{07}$ (1.0×) \\
&    FedAU        & 48.3$_{\pm 2.3}$ & 35.6 (5.7$_{\pm 2.4}$) & 45.6 (6.4$_{\pm 3.6}$) & 33.6 (5.7$_{\pm 4.1}$) & 7.90e$^{09}$ (18.2×) & 6.60e$^{13}$ (18.2×) & 4.49e$^{07}$ (1.0×) \\
& NoT & 48.1$_{\pm 2.2}$ & 39.2 (9.1$_{\pm 2.8}$) & 50.5 (11.3$_{\pm 3.4}$) & 41.8 (13.9$_{\pm 4.5}$) & 4.74e$^{09}$ (30.4×) & 3.96e$^{13}$ (30.3×) & 4.49e$^{07}$ (1.0×) \\
& PUF-Special  & 48.5$_{\pm 2.1}$ & 34.1 (4.0$_{\pm 3.7}$) & 46.1 (6.6$_{\pm 3.5}$) & 33.5 (5.6$_{\pm 4.2}$) & \textbf{2.48e$^{09}$ (58.1×)} & \textbf{2.07e$^{13}$ (58.0×)} & 4.49e$^{07}$ (1.0×) \\
& PUF-Regular  & 47.9$_{\pm 2.3}$ & \textbf{31.2} (\textbf{2.9}$_{\pm 1.9}$) & \textbf{42.9} (\textbf{3.9}$_{\pm 2.9}$) & \textbf{31.4} (\textbf{3.8}$_{\pm 3.1}$) & 7.54e$^{09}$ (19.1×) & 5.55e$^{13}$ (21.6×) & 4.49e$^{07}$ (1.0×) \\
\midrule

\multirow{6}{*}{\makecell[l]{\textbf{Multiple Unlearning},\\ CIFAR-100, Non-IID,\\ MiT-B0, $E$=1}}
& Original     & 75.0$_{\pm 0.0}$ & 84.6$_{\pm 5.3}$ & 76.8$_{\pm 9.0}$ & 73.8$_{\pm 1.4}$ & --- & --- & --- \\
& Retrain      & 70.9$_{\pm 1.3}$ & 54.3$_{\pm 3.5}$ & 45.4$_{\pm 3.6}$ & 43.5$_{\pm 1.6}$ & 1.06e$^{10}$ (1.0×) & 3.40e$^{15}$ (1.0×) & 1.33e$^{07}$ (1.0×) \\
&    FedAU        & 71.4$_{\pm 0.9}$ & 58.9 (4.6$_{\pm 2.3}$) & 50.1 (4.7$_{\pm 2.3}$) & 47.2 (3.7$_{\pm 1.9}$) & 3.19e$^{09}$ (3.3×) & 1.02e$^{15}$ (3.3×) & 1.33e$^{07}$ (1.0×) \\
& NoT & 71.4$_{\pm 0.9}$ & 69.9 (15.6$_{\pm 6.3}$) & 62.5 (16.6$_{\pm 8.9}$) & 59.0 (15.5$_{\pm 7.5}$) & \textbf{6.80e$^{08}$ (15.6×)} & \textbf{2.18e$^{14}$ (15.6×)} & 1.33e$^{07}$ (1.0×) \\
& PUF-Special  & 71.3$_{\pm 1.4}$ & 60.6 (6.3$_{\pm 2.1}$) & 53.0 (7.3$_{\pm 2.7}$) & 50.5 (7.0$_{\pm 1.8}$) & 1.12e$^{09}$ (9.5×) & 3.57e$^{14}$ (9.5×) & 1.33e$^{07}$ (1.0×) \\
& PUF-Regular  & 70.9$_{\pm 1.5}$ & \textbf{56.6} (\textbf{2.3}$_{\pm 1.1}$) & \textbf{48.7} (\textbf{2.6}$_{\pm 2.3}$) & \textbf{45.6} (\textbf{2.1}$_{\pm 1.1}$) & 3.93e$^{09}$ (2.7×) & 1.17e$^{15}$ (2.9×) & 1.33e$^{07}$ (1.0×) \\
\bottomrule
\end{tabular}
}
\caption{\major{Performance of PUF and other baselines across different settings for multiple unlearning when two clients request forgetting. For efficacy metrics, baseline results are expressed as \textit{mean metric value (mean $\Delta$ $\pm$ standard deviation)}, with $\Delta$ in parentheses representing the average absolute difference from the Retrain baseline. Lower $\Delta$ corresponds to better unlearning performances. For efficiency metrics, higher $\times$ (reduction over Retrain baseline) are better. The best-performing method is shown in bold.}}
\label{table:multiple_unlearning_after_reviews}
\end{table*}
 \captionsetup{labelfont={color=black}}

%% file: tables/segmentation_table.tex
\begin{table*}[!t]
\small
\centering
\adjustbox{max width=0.8\textwidth}{
\begin{tabular}{llccccc}
\toprule
\multirow{3}{*}{\makecell[l]{\textbf{Setting}}} & \multirow{3}{*}{\makecell[l]{\textbf{Method}}} & \multicolumn{3}{c}{\textbf{Efficacy}} & \multicolumn{2}{c}{\textbf{Efficiency}} \\
\cmidrule(lr){3-5} \cmidrule(lr){6-7}
& & \textbf{Test Acc.} & \textbf{Forget Acc. ($\Delta$ $\downarrow$)} & \textbf{Forget Loss ($\Delta$ $\downarrow$)} & \textbf{Communication} & \textbf{Computation} \\
& & & & & Bytes ($\times$ $\uparrow$) & FLOPs ($\times$ $\uparrow$) \\
\midrule

\multirow{5}{*}{\makecell[l]{\textbf{ProstateMRI},\\ U-NET}}
& Original     & 86.5$_{\pm 0.0}$ & 88.9$_{\pm 4.6}$ & 0.08$_{\pm 0.05}$ & --- & --- \\
& Retrain      & 82.7$_{\pm 1.7}$ & 73.3$_{\pm 7.2}$ & 0.22$_{\pm 0.10}$ & 2.74e$^{11}$ (1.0×) & 5.49e$^{16}$ (1.0×) \\
& PGA          & 82.9$_{\pm 0.4}$ & 77.0 (3.8$_{\pm 2.3}$) & 0.18 (0.06$_{\pm 0.02}$) & 2.69e$^{10}$ (10.2×) & 5.44e$^{15}$ (10.1×) \\
& NoT            & 83.4$_{\pm 0.7}$ & 77.4 (3.9$_{\pm 2.4}$) & 0.17 (0.06$_{\pm 0.02}$) & 2.63e$^{10}$ (10.4×) & 5.27e$^{15}$ (10.4×) \\
& PUF-Special  & 83.1$_{\pm 0.4}$ & \textbf{76.7} (\textbf{3.4}$_{\pm 2.3}$) & \textbf{0.17} (\textbf{0.05}$_{\pm 0.02}$) & \textbf{2.30e$^{10}$ (11.9×)} & \textbf{4.62e$^{15}$ (11.9×)} \\
\bottomrule
\end{tabular}
}
\caption{\major{Performance of PUF and other baselines across for ProstateMRI segmentation task. For efficacy metrics, Lower $\Delta$ corresponds to better unlearning performances. For efficiency metrics, higher $\times$ (reduction over Retrain baseline) are better. The best-performing method is shown in bold.}}
\label{table:prostate_unet_after_review}
\end{table*}
 \captionsetup{labelfont={color=black}} 

%% file: figures/hp_tuning_figures.tex
\begin{figure*}[t!]{
\centering
\begin{subfigure}[t]{0.32\textwidth} 
    \centering
    \includegraphics[width=0.95\linewidth]{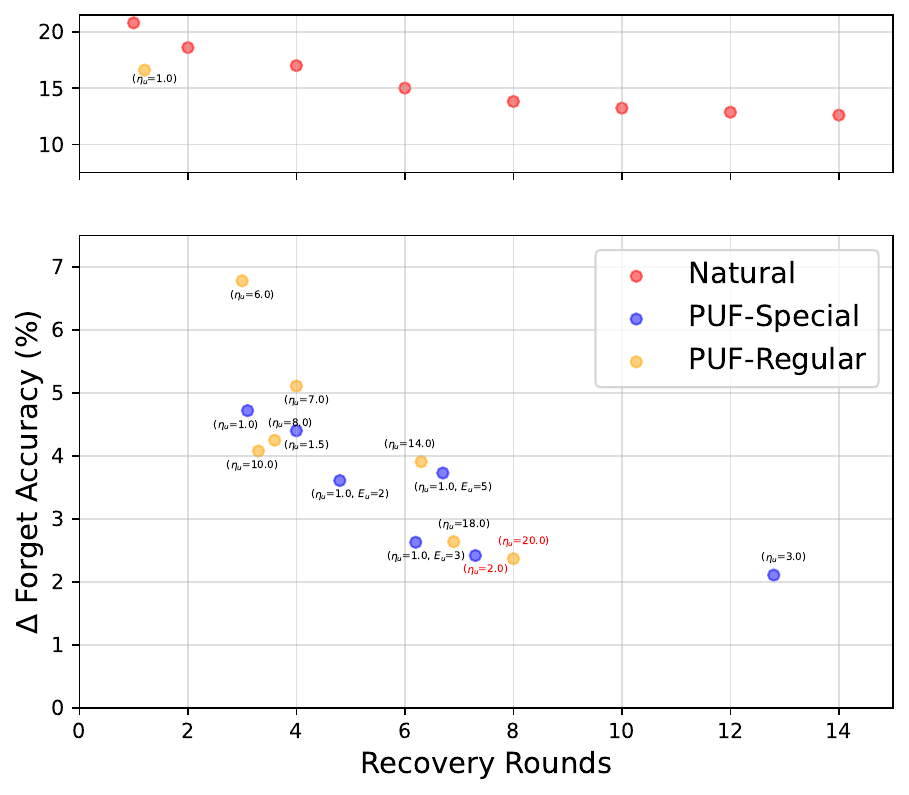}
    \caption{CIFAR-100, Non-IID, ResNet18.}
    \label{fig:hp_cifar100}
\end{subfigure}
\hfill
\begin{subfigure}[t]{0.32\textwidth}
    \centering
    \includegraphics[width=0.95\linewidth]{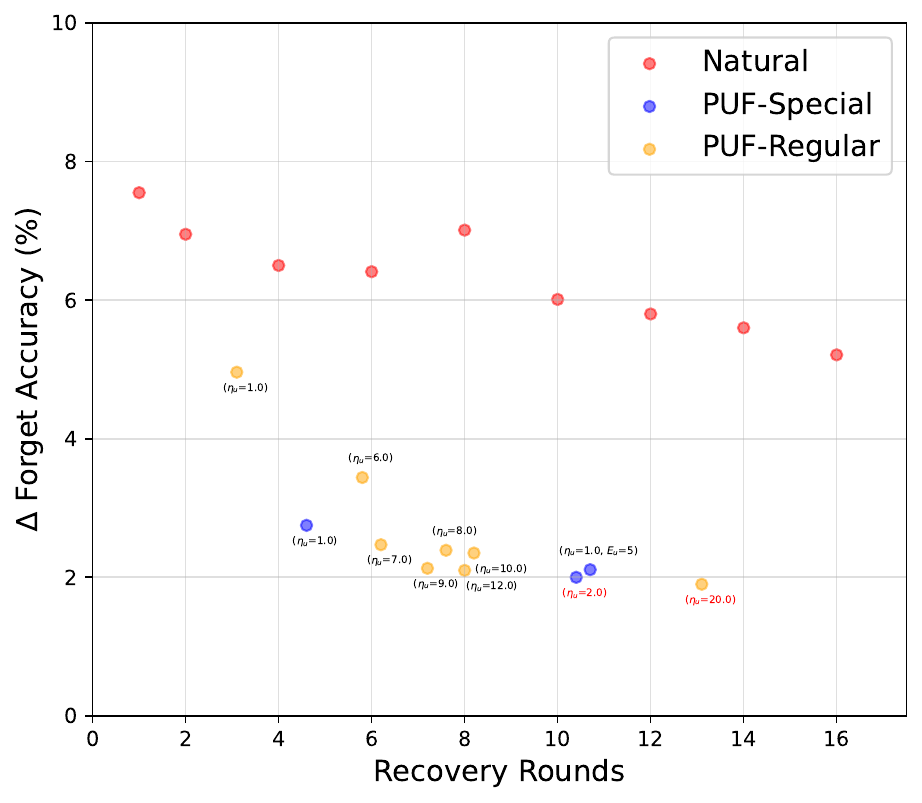}
    \caption{CIFAR-10, Non-IID, ResNet18.}
    \label{fig:hp_cifar10}
\end{subfigure}
\hfill
\begin{subfigure}[t]{0.32\textwidth}
    \centering
    \includegraphics[width=0.95\linewidth]{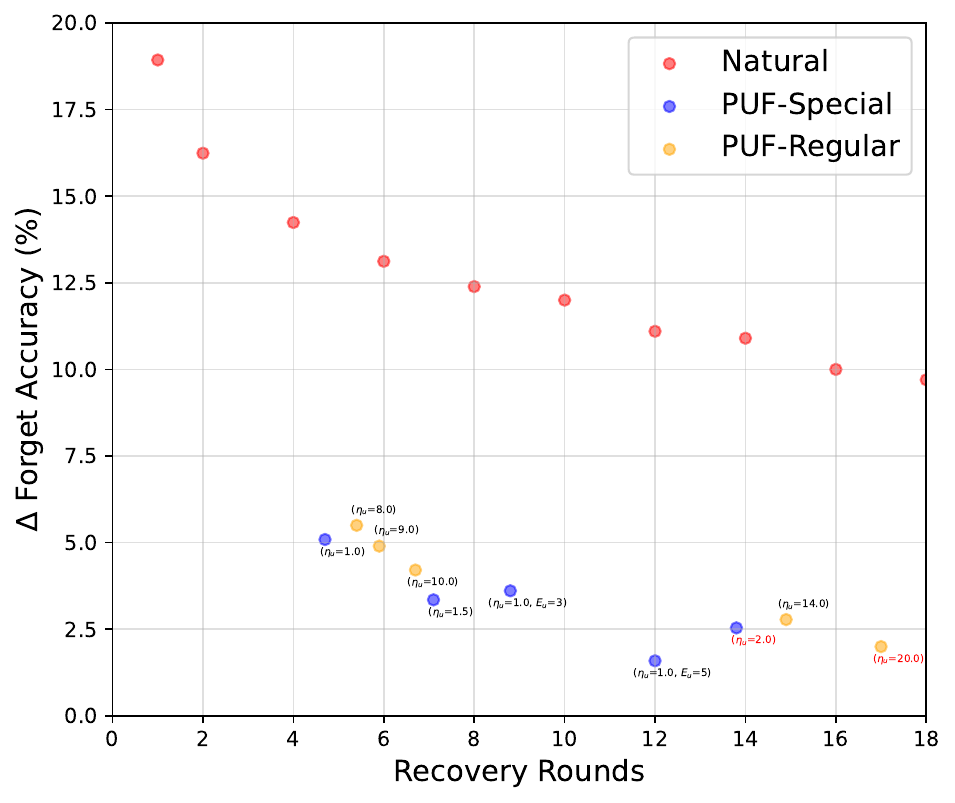}
    \caption{CIFAR-100, Non-IID, MiT-B0.}
    \label{fig:hp_mitb0}
\end{subfigure}
\caption{\textbf{Performance of PUF with varying hyper-parameters} ($\eta_u$, $E_u$). When $E_u$ is omitted, it is set to 1. \textbf{X-axis}: number of recovery rounds required to match the retrained model's test accuracy. \textbf{Y-axis}: gap in forget accuracy compared to the retrained model. Points closer to the origin indicate better performance. The experimental setting is indicated in subcaption as a triple \textit{dataset, data distribution, model architecture}. The \textit{Natural} baseline reports results for a naive strategy of fine tuning the global model without the participation of the unlearning client, no explicit unlearning is performed.}
\label{fig:hp_tuning_puf}
}
\end{figure*}

%% file: figures/hp_tuning_segmentation.tex
\begin{figure}[!t]
\centering
\includegraphics[width=0.75\columnwidth]{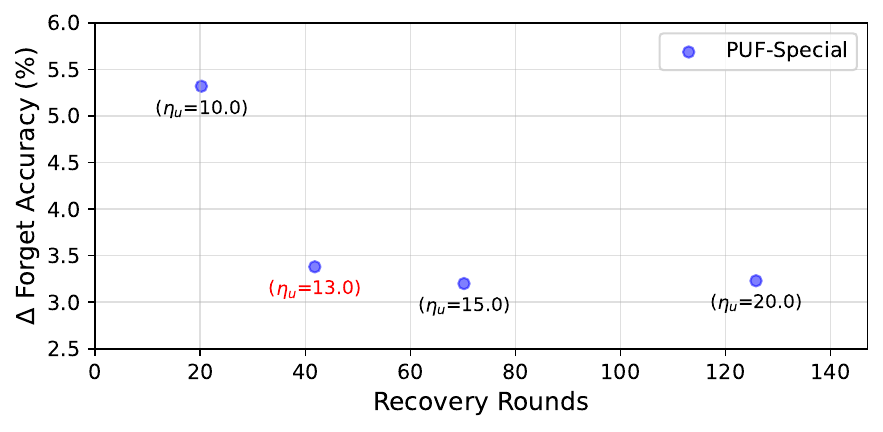}
\caption{\textbf{Performance of PUF with varying hyper-parameter} $\eta_u$ for the ProstateMRI setting.}
\label{fig:hp_segm}
\end{figure} 

%% file: sections/conclusion.tex
\section{Conclusion}
\label{sec:conclusion}
In this paper, we introduced \tool{}, a novel FU method that leverages negated pseudo-gradients to enable clients to exercise their right to be forgotten. Aligned with the design principles of FL, \tool{} does not require storing any processed client information and relies only on ephemeral updates, even during unlearning. It integrates seamlessly with standard FedAvg, by requiring no significant modifications or additional computational/communication overhead. As a result, \tool{} is both task-agnostic and capable of handling multiple unlearning requests concurrently, which are two critical aspects that most existing literature does not consider. Experimental results on two classification datasets and a segmentation task, across varying degrees of data heterogeneity and different model architectures, are reported to show the general applicability of the proposed approach. \tool{} has clearly demonstrated to be more efficient in recovering expected performance and more accurate in inducing forgetting of requested data compared to state-of-the-art baselines.

%% file: sections/appendix.tex
\newpage
\newpage 

\appendices
\section{Hyper-parameter Tuning}
\label{appendix:hp_tuning}
We compared \tool{} against five state-of-the-art baseline methods: FedEraser [16], PGA [19], FedAU [21], MoDe [20], and NoT [22]. Below, we summarize the hyper-parameter tuning for each method.

\smallskip
\noindent \textbf{FedEraser} [16]: we set the retention interval of rounds between stored updates to 1 (tuned in $\{1, 2\}$) and the calibration epochs (number of local epochs during calibration training) to 0.5.

\smallskip
\noindent \textbf{PGA} [19]: we adopted the same hyper-parameters prescribed in the original paper, i.e., gradient clipping set to $5.0$, batch size in $\{512, 1024\}$, local unlearning epochs set to $5$, and projection on the $\ell_2$-ball of radius $\delta$ around $w_{\text{ref}}$. We additionally tuned their early stopping threshold in the range $[2.2, 10.0]$, selecting $9.0$ for the IID setting and $6.5$ for the Non-IID setting. 

\smallskip
\noindent \textbf{MoDe} [20]: we set the number of memory-guidance rounds to $10$ and degradation rounds to $6$, the learning rate for memory guidance to $0.0005$, and the momentum parameter to $0.95$ (tuned in the range $[0.9, 0.99]$). 

\smallskip
\noindent \textbf{FedAU} [21]: we used $10$ local epochs to train the auxiliary module, and tuned the weight for auxiliary module integration in $[0.0, 1.0]$, selecting $0.04$ as the best configuration. 

\smallskip
\noindent \textbf{NoT} [22]: we negated the weights of the first layer of the neural network, following the prescription in the original paper.

\smallskip
\noindent \textbf{\tool{}}: for PUF-Regular, we set the scaling factor for the aggregated update from unlearning clients to $\eta_u = 20.0$ and the learning rate for retained clients to $\eta_r = 1.0$. For PUF-Special, we set $\eta_u = 2.0$. Unless otherwise specified, we used SGD as the server-side optimizer with $\eta_s = 1.0$. 

\section{Cost Estimation Methodology}
\label{appendix:cost_estimation}
We report three complementary efficiency metrics: communication cost (in bytes), computational cost (in FLOPs), and storage overhead (in bytes).  
These metrics, preferred over wall‐clock time as in [22], are hardware- and implementation-agnostic, enabling fair comparisons across tasks, datasets, and systems.  
Unlike time, they isolate algorithmic efficiency from factors such as hardware heterogeneity or parallelization.  
We also report the relative improvement over the Retrain baseline.  
For each method, we account for both the unlearning phase and the recovery phase, with the latter corresponding to standard FedAvg rounds among \(C_r\) clients.

\begin{table}[!t]
\centering
\setlength{\tabcolsep}{4pt}
\renewcommand{\arraystretch}{1.05}
\begin{tabular}{l@{\hspace{6pt}}l}
\hline
\textbf{Symbol} & \textbf{Description} \\
\hline
\(B\) & Bytes per parameter (4 for \texttt{float32}) \\
\(P\) & Model parameters \\
\(P_c\) & Classifier params (FedAU) \\
\(C\) & Total clients \\
\(C_u\) & Unlearning clients  \\
\(C_r\) & Remaining clients \(= C - C_u\) \\
\(F\) & FLOPs per image (\(0.15\)G ResNet-18, \(1.7\)G MiT-B0) \\
\(N\) & Samples per client \\
\(E\) & Local epochs (generic) \\
\(R_\mathrm{ret}\) & Retained/calibration rounds (FedEraser) \\
\(E_\mathrm{cal}\) & Epochs per calibration round (FedEraser) \\
\(E_\mathrm{asc}\) & Epochs of ascent (PGA) \\
\(R_d, R_m\) & Degradation and memory-guidance rounds (MoDe) \\
\(R_\mathrm{rec}\) & Recovery rounds \\
\hline
\end{tabular}
\caption{{\color{black}Notation used for cost estimation formulas.}}
\label{tab:notation}
\end{table}
\subsection{Communication Cost}
For each round, we compute the total bytes exchanged across participating clients (uplink + downlink) in both the unlearning phase and the recovery phase.  
The overall communication cost is obtained by summing the costs of these two phases.
The general per‐round communication cost is:
\[
\mathrm{Comm}_t =
2 \, P \, B \, |\mathcal{S}_t|
\]
where \( |\mathcal{S}_t| \) is the number of clients participating in round \(t\).

For the recovery phase (same for all methods), the communication cost is:
\[
\mathrm{Comm}_{\mathrm{rec}} =
2 \, P \, B \, C_r \, R_{\mathrm{rec}}
\]

For the unlearning phase, the communication costs are method-specific, as reported in Table~\ref{tab:cost_formulas}.

\subsection{Computational Cost}
The computational cost (in FLOPs) includes both the unlearning phase and the recovery phase.  
We generically estimate FLOPs in a training round as:
\[
\mathrm{FLOPs} = F \, N \, E \, C
\]
with \(F\) denoting FLOPs per image.

For the \textbf{recovery phase}, the computational cost is:
\[
\mathrm{Comp}_{\mathrm{rec}} = F \, N \, E \, C_r \, R_{\mathrm{rec}}
\]

For the \textbf{unlearning phase}, the computational costs per method are shown in Table~\ref{tab:cost_formulas}.

\subsection{Storage Overhead}
We report the maximum persistent storage (in bytes) required across clients and/or the server over multiple rounds and necessary for unlearning.  
The baseline cost for storing one global model is:
\[
\mathrm{Storage}_{\mathrm{baseline}} = P \, B
\]
which corresponds to the Retrain baseline following FedAvg.  
Storage requirements per method are summarized in Table~\ref{tab:cost_formulas}.

\begin{table}[!t]
\centering
\setlength{\tabcolsep}{6pt}
\renewcommand{\arraystretch}{1.15}
\begin{adjustbox}{max width=\columnwidth}
\begin{tabular}{lccc}
\hline
\textbf{Method} & \textbf{Communication Cost} & \textbf{Computation Cost} & \textbf{Storage Overhead} \\
\hline
FedEraser & \( 2 \, P \, B \, C_r \, R_{\mathrm{ret}} \) & \( F \, N \, E_{\mathrm{cal}} \, C_r \, R_{\mathrm{ret}} \) & \( C \, R_{\mathrm{ret}} \, P \, B \) \\
PGA & \( 2 \, P \, B \) & \( F \, N \, E_{\mathrm{asc}} \, C_u \) & \( C \, P \, B \) \\
FedAU & \( 2 \, P_c \, B \) & Negligible & \( P \, B \) \\
MoDe & \( 2 \, P \, B \, \big(C_r \, R_d + C \, R_m\big) \) & \( F \, N \, E \, \big(C_r \, R_d + C \, R_m\big) \) & \( 2 \, P \, B \) \\
PUF-Special & \( 2 \, P \, B \, C_u \) & \( F \, N \, E \, C_u \) & \( P \, B \) \\
PUF-Regular & \( 2 \, P \, B \, C \) & \( F \, N \, E \, C \) & \( P \, B \) \\
NoT & Negligible & Negligible & \( P \, B \)\\
\hline
\end{tabular}
\end{adjustbox}
\caption{Communication, computation, and storage cost formulas for each unlearning method. Notation described in Table \ref{tab:notation}.}
\label{tab:cost_formulas}
\end{table}